\documentclass[final]{nesy2025} 

\usepackage{longtable}

\usepackage{booktabs}
\usepackage[load-configurations=version-1]{siunitx} 
\usepackage[most]{tcolorbox}
\usepackage{float}
\usepackage{multirow}
\usepackage{bbm}
\usepackage{enumitem}
\usepackage{bm}
\usepackage{wrapfig}

\lstdefinestyle{mystyle}{
basicstyle=\ttfamily\footnotesize,
captionpos=t,
belowcaptionskip=0pt,
frame=lines,
framerule=1pt,
}
\newcommand{\aspnot}{\text{not}}
\newcommand{\indicatorf}{\mathbbm{1}}
\newcommand{\sign}{\text{sign}}

\SetKw{Continue}{continue}

\tcbset{
  example/.style={
    enhanced,
    colback=gray!10,
    colframe=gray!50,
    coltitle=black,
    fonttitle=\small\bfseries,
    fontupper=\scriptsize,
    attach boxed title to top left={yshift=-\tcboxedtitleheight/2},
    boxed title style={
      colback=gray!50,
      size=small,
      leftrule=0pt,
      rightrule=0pt,
      toprule=0pt,
      bottomrule=0pt
    }
  }
}
\newtcbtheorem[number within=section]{example-box}{Example}{example}{ex}

\title[Disentangling Neural Disjunctive Normal Form Models]{Disentangling
Neural Disjunctive Normal Form Models}

\author{\Name{Kexin Gu Baugh} \Email{kexin.gu17@imperial.ac.uk}\\
\addr Imperial College London, UK
\AND
\Name {Vincent Perreault} \Email{vincent.perreault@polymtl.ca} \\
\addr Polytechnique Montréal, Canada
\AND
\Name{Matthew Baugh} \Email{matthew.baugh17@imperial.ac.uk}\\
\addr Imperial College London, UK
\AND
\Name{Luke Dickens} \Email{l.dickens@ucl.ac.uk}\\
\addr University College London, UK
\AND
\Name{Katsumi Inoue} \Email{inoue@nii.ac.jp}\\
\addr National Institute of Informatics, Japan
\AND
\Name{Alessandra Russo} \Email{a.russo@imperial.ac.uk}\\
\addr Imperial College London, UK
}

\begin{document}

\maketitle

\begin{abstract}
    Neural Disjunctive Normal Form (DNF) based models are powerful and
    interpretable approaches to neuro-symbolic learning and have shown promising
    results in classification and reinforcement learning settings without prior
    knowledge of the tasks.
    However, their performance is degraded by the thresholding of the
    post-training symbolic translation process.
    We show here that part of the performance degradation during translation is
    due to its failure to disentangle the learned knowledge represented in the
    form of the networks' weights.
    We address this issue by proposing a new disentanglement method; by
    splitting nodes that encode nested rules into smaller independent nodes, we
    are able to better preserve the models' performance.
    Through experiments on binary, multiclass, and multilabel classification
    tasks (including those requiring predicate invention), we demonstrate that
    our disentanglement method provides compact and interpretable logical
    representations for the neural DNF-based models, with performance closer to
    that of their pre-translation counterparts. Our code is available at
    \url{https://github.com/kittykg/disentangling-ndnf-classification}.
\end{abstract}

\section{Introduction} \label{sec:intro}

Neuro-symbolic learning methods integrate neural models' learning capabilities
with symbolic models' interpretability. These methods have expanded beyond
differentiable inductive logic programming \citep{delta-ilp,hri}, showing
success in reinforcement learning \citep{nsrl-fol,nudge,ndnf-mt}, transition
systems \citep{dlfit, delta-lfit-2}, and classification
\citep{pix2rule,ns-classifications}. Their rule-based interpretations offer more
faithful and reliable model explanations compared to post-hoc approaches
\citep{interpretable-model}.

Many neuro-symbolic learning methods build upon inductive logic programming
(ILP) \citep{ilp, ilasp}, requiring background knowledge for initial clauses
\citep{nesy-pi}, rule templates \citep{delta-ilp,dilp-structured,hri,lnn-ilp},
mode biases \citep{ns-asp-raw-data,nudge}, or search space constraints
\citep{alpha-ilp}. While the ability to use background knowledge is a merit
inherited from ILP, the hypothesis search space can be too large for the model
to handle when prior knowledge is unavailable. And if the background knowledge
injected by the user is incorrect, an optimal solution may be excluded from the
search space. Additionally, these methods typically require predicate-based
inputs, necessitating either manual definition \citep{nudge} or extraction using
pre-trained components \citep{nsrl-fol,ffnsl,alpha-ilp,nesy-pi} when the inputs
are not logical or structured. In contrast, neural methods such as MLPs can be
trained without background knowledge, regardless of the input format, trading
interpretability for flexibility. Decision trees are good alternatives to MLPs
when interpretability is more important, but the trees can grow too large to be
readable. The Optimal Sparse Decision Tree (OSDT) \citep{osdt} addresses tree
complexity but is limited to binary features and lacks support for multiclass
classification. On the other hand, the recently developed neural DNF-based model
\citep{pix2rule,ns-classifications,ndnf-mt} represents a promising amalgamation
of interpretability and flexibility, and operates without the need for
task-specific background knowledge. Like MLPs, the neural DNF-based models are
end-to-end differentiable, but under appropriate conditions can be translated to
formal logic programmes. Unlike the mentioned neuro-symbolic approaches, the
neural DNF-based models offer a broader space of possible rules and seamlessly
integrate with other neural components \citep{pix2rule,ndnf-mt}. However,
performance can degrade when converted from neural to logical form, a critical
step in providing interpretations for trained models.

In this paper, we make three major contributions.
(1) We identify that the existing thresholding discretisation method fails to
faithfully translate some neural units into logical form, causing a performance
drop, and characterise such neural units as \emph{entangled}.
(2) We propose a disentanglement method to address this issue, better preserving
model performance in the extracted symbolic representation.
(3) We introduce a simple threshold-learning method for real-valued features.
These learned bounds are of the form `$\mathit{feature\ >\ learned\ threshold\
value}$' and act as interpretable invented predicates.
We demonstrate that our proposed disentanglement method drastically decreases
the performance degradation when translating trained neural DNF-based models
across binary, multiclass, and multilabel classification tasks (some of which
require predicate invention).

\section{Entangled Nodes in Neural DNF-based Models}

A neural DNF model \citep{pix2rule} is built up of \emph{semi-symbolic nodes}
that behave as soft conjunctions or disjunctions. Such a model is constructed
with a layer of conjunctive nodes followed by a layer of disjunctive nodes. A
semi-symbolic node with trainable weights $w_i$, $i = 1, \ldots, N$ and
parameter $\delta$, under input $\mathbf{x}$ gives output:
\begin{align} \label{eq:ss-node}
    \hat{y} & = \tanh(f_{\mathbf{w}}(\mathbf{x}))
    = \tanh\left(\sum_{i=1}^{N} w_i x_i + \beta\right),
    \; \text{with} \;
    \beta   = \delta \left(\max_{i}|w_i| - \sum_{i=1}^{N} |w_i|\right)
\end{align}
where $x_i \in [-1, 1]$, with the extreme value $1$ ($-1$) interpreted as $\top$
($\bot$) and intermediate values interpreted as degrees of beliefs. Activation
$\hat{y}$ is interpreted similarly, but cannot take extreme values $\pm 1$. A
bivalent interpretation of $\hat{y}$ treats $\hat{y} > 0$ ($\le 0$) as $\top$
($\bot$). $\delta$ induces behaviour similar to a conjunction (disjunction) when
$\delta = 1$ ($-1$).\footnote{See \appendixref{apd:additional-background} for a
more detailed background introduction.}

We introduce the \textbf{soft-valued truth table} to describe the behaviour of a
semi-symbolic node, as shown in Example~\ref{ex:soft-truth-table}.

\begin{definition}[Soft-valued Truth Table of Semi-symbolic Node]\label{def:soft-truth-table}
    A soft-valued truth table captures the output behaviour of a semi-symbolic
    node with weights $\mathbf{w} \in \mathbb{R}^N$ over all possible bivalent
    inputs $\mathcal{X} = \{-1,1\}^N$. Each row starts with a possible
    combination of input values $\mathbf{x} \in \mathcal{X}$, followed by the
    corresponding node's activation $\tanh (f_{\mathbf{w}}(\mathbf{x}))$ and
    bivalent interpretation $b = \top$ if $\tanh (f_{\mathbf{w}}(\mathbf{x})) >
    0$ and $\bot$ otherwise.
\end{definition}

\begin{example-box}{Soft-valued truth table}{soft-truth-table}

Consider a trained conjunctive semi-symbolic node with weights $\mathbf{w} =
[-6, -2, -2, 2, -6]$. Its bias is calculated as $\beta = \max |\mathbf{w}| -
\sum |\mathbf{w}| = -12$. A soft-valued truth table of this node is shown below,
with some rows with $b = \bot$ omitted for brevity.
\begin{table}[H]
    \centering
    \scriptsize
    \begin{tabular}{ccccccc}
        \toprule
        \multirow{2}{*}{$x_1$} & \multirow{2}{*}{$x_2$} & \multirow{2}{*}{$x_3$} & \multirow{2}{*}{$x_4$} & \multirow{2}{*}{$x_5$} & Node activation                         & Bivalent Interpretation \\
                               &                        &                        &                        &                        & $\tanh(\sum_{i=1}^{5} w_i x_i + \beta)$ & $b$                     \\ \midrule
        1                      & 1                      & 1                      & 1                      & 1                      & -1.000                                  & $\bot$                  \\
        \multicolumn{7}{c}{...}                                                                                                                                                                        \\
        -1                     & 1                      & -1                     & 1                      & -1                     & 0.964                                   & $\top$                  \\
        -1                     & -1                     & 1                      & 1                      & -1                     & 0.964                                   & $\top$                  \\
        -1                     & -1                     & -1                     & 1                      & -1                     & 1.000                                   & $\top$                  \\
        -1                     & -1                     & -1                     & -1                     & -1                     & 0.964                                   & $\top$                  \\
        \bottomrule
    \end{tabular}
\end{table}

\end{example-box}

A neural DNF-based model can be translated to a symbolic representation via an
automatic post-training process \citep{pix2rule,ndnf-mt}. One key step in the
post-training process is to convert the weights from a continuous range to a
fixed-valued set $\{-6, 0, 6\}^N$ ($\pm$ 6 to saturate $\tanh$), so that the
node can be translated into a bivalent logic representation with provable
truth-value equivalence \citep{ndnf-mt}. We call this the
\textbf{discretisation} process. The discretisation method used in mentioned
previous works is the thresholding method: it selects a value $\tau$ to filter
out weights with small absolute values while maintaining the model's performance
as much as possible:
\begin{equation*}
    g_{\text{threshold}}(\mathbf{w} ; \tau) = \hat{\mathbf{w}},
    \; \text{with} \;
    \hat{w_i} = 6 \cdot \sign (w_i) \cdot \indicatorf(|w_i| > \tau)
\end{equation*}
where $\mathbbm{1} ( \alpha ) = 1$ if condition $\alpha$ is true and 0 otherwise
and $0 \le \tau \le \max |\mathbf{w}|$. Thresholding is usually applied on the
whole neural DNF-based model with a shared threshold $\tau$ across all
conjunctive and disjunctive nodes \citep{pix2rule, ns-classifications}.
\citet{ndnf-mt} only thresholds the conjunctive nodes so that the conjunctive
layer is translated into bivalent rules while translating the disjunctive layer
into probabilistic rules. To encourage the weights to be in $\{-6, 0, 6\}^N$,
\citet{ndnf-mt} uses an auxiliary loss function $|w||6 - |w||$ during training.
However, due to the flexibility of the neural architecture, a semi-symbolic node
is more likely to learn an \textbf{entangled} representation that cannot be
translated into a single logical formula.

Example~\ref{ex:soft-truth-table} shows a conjunctive node with entangled
weights. The discretisation that best preserves the node's behaviour uses $\tau
= 0$, resulting in a discretised conjunctive node with weights $[-6, -6, -6, 6,
-6]$ and is translated into: $\text{c} \leftarrow \aspnot\ \text{a}_1,\ \aspnot\
\text{a}_2,\ \aspnot\ \text{a}_3,\ \text{a}_4,\ \aspnot\ \text{a}_5$. Although
this is the best translation, it still alters the behaviour slightly: rule head
$\text{c}$ cannot be true when $\text{a}_4$ is not true, but it is possible for
the original node's activation to be interpreted as true when $x_4 = -1$. This
is because the thresholding method overemphasises the smaller magnitude weights
$w_2, w_3,$ and $w_4$. In fact, looking at the soft-valued truth table, the
behaviour can be perfectly described by a set of three rules:
\begin{align*}
    \mathfrak{L} = \left\{
        \begin{array}{l}
            \text{c} \leftarrow \aspnot\ \text{a}_1,\ \aspnot\ \text{a}_3,\ \text{a}_4,\ \aspnot\ \text{a}_5.          \\
            \text{c} \leftarrow \aspnot\ \text{a}_1,\ \aspnot\ \text{a}_2,\ \aspnot\ \text{a}_3,\ \aspnot\ \text{a}_5. \\
            \text{c} \leftarrow \aspnot\ \text{a}_1,\ \aspnot\ \text{a}_2,\ \text{a}_4,\ \aspnot\ \text{a}_5.
        \end{array}
    \right\}.
\end{align*}
By the lower-weighted terms only appearing in subsets of the rule bodies, it
reflects the smaller influence of $x_2, x_3, x_4$ on the node's behaviour. This
phenomenon often occurs in trained neural DNF-based models, despite having the
auxiliary loss function $|w||6 - |w||$ to penalise such behaviour.

\section{Disentanglement of Semi-symbolic Node} \label{sec:disentangle-conj}
\begin{figure}[!ht]
    \centering
    \includegraphics[width=0.95\textwidth]{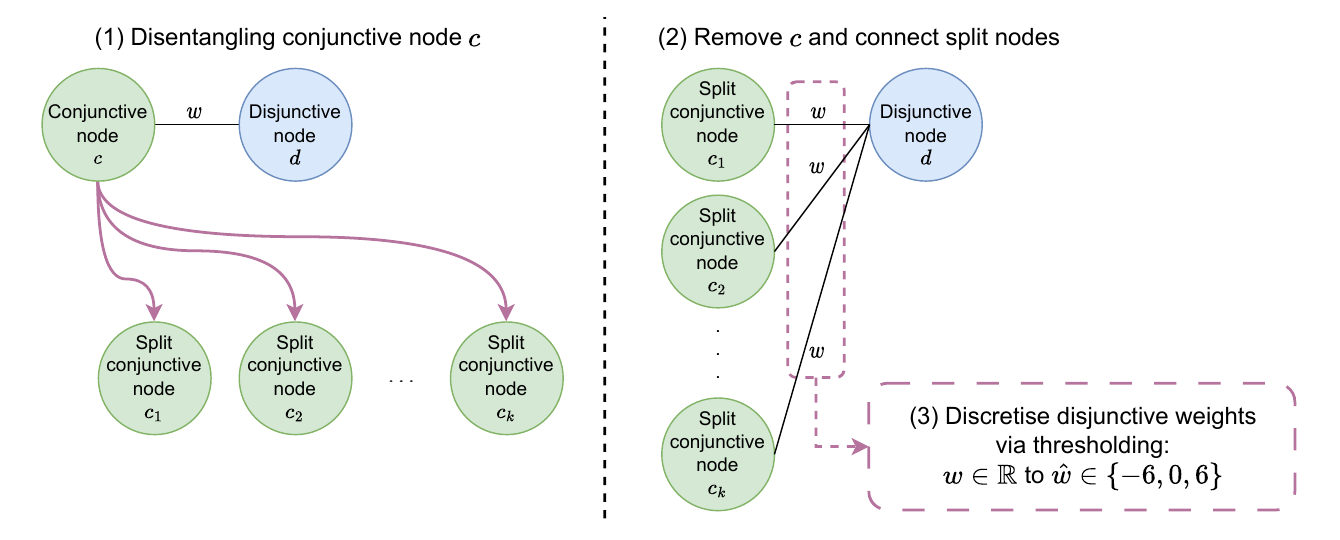}
    \vspace{-1em}
    \caption{The three steps of the disentanglement method. Here $w$ is the
        weight connecting conjunctive node $c$ to disjunctive node $d$.}
    \label{fig:disentangle-conj}
    \vspace{-1.5em}
\end{figure}

We propose a method that \textbf{disentangles} the weights of semi-symbolic
nodes while preserving nodes' behaviour. The approach transforms a real-valued
conjunctive node into multiple smaller discretised nodes that jointly represent
the same behaviour as the original node's. The process consists of three steps
(\figureref{fig:disentangle-conj}): (1) split all conjunctive nodes into smaller
disentangled and discretised conjunctive nodes; (2) replace the original
conjunctive node with the split conjunctive nodes by connecting them to the
disjunctive node with the same weight as the original node; and (3) if the model
is not a neural DNF-MT model\footnote{Neural DNF-MT model proposed by
\cite{ndnf-mt} is designed to output probabilities.}, apply the thresholding
discretisation method on the disjunctive layer (either on 0 or sweeping through
all possible values). We only disentangle conjunctive nodes while still applying
thresholding to discretise the disjunctive layer because:
\begin{enumerate}
    \item Split conjunctive nodes can be reconnected to the neural DNF-based
          model under logical equivalence without auxiliary predicates. For
          instance, if a conjunctive node $c$, connected to a disjunctive node
          $d$ with a positive weight, splits into $c_1, \ldots, c_k$, then $(c_1
              \lor \ldots \lor c_k) \rightarrow d \equiv (c_1 \rightarrow d) \land
              \ldots \land (c_k \rightarrow d)$.

    \item Step (1) of our method requires binary $\{-1, 1\}^N$ inputs. While
          conjunctive layer inputs are typically binary, disjunctive layer
          inputs cannot be guaranteed to be binary, even with auxiliary loss
          functions.
\end{enumerate}

\subsection{Disentangling a Conjunctive Node} \label{sec:smaller-split}
To disentangle a conjunctive node, the goal is to find a collection of new
conjunctive nodes that jointly encode its interpreted behaviour. To this end, we
can make use of the following:
\setcounter{theorem}{0}
\begin{proposition}
    Given a conjunctive node $c$ with $s$ inputs parameterised by $w \in
        \mathbb{R}^s$, there exists a collection of conjunctive nodes $\{c_i\}$
        with associated (split) weights $\{ \tilde{\mathbf{w}}_i \} =
        \mathcal{W}$, where each $\tilde{\mathbf{w}}_i$ is discretised, i.e.
        $\tilde{\mathbf{w}}_i \in \{-6, 0, 6\}^s$, such that under discrete
        input $\{-1, 1\}^s$ the disjunction of their bivalent interpretations is
        the same as that of $c$.
\end{proposition}

In this section we focus on disentangling a conjunctive node $c$ connected to a
disjunctive node with \emph{a positive weight}.\footnote{For the negative weight
case, see \appendixref{apd:neg-conj-disentangle}.}

Using \equationref{eq:ss-node}, the raw output of $c$ for input $\mathbf{x} \in
\{-1, 1\}^{s}$ can be written as:
\begin{equation} \label{eq:conjuncitve-node-raw-output}
    f_{\mathbf{w}}(\mathbf{x}) =
    \sum_{j \in \mathcal{J}} w_j x_j + \max_{j \in \mathcal{J}} |w_j| - \sum_{j \in \mathcal{J}} |w_j|
\end{equation}
where the relevant set of indices of non-zero weights $\mathcal{J} = \left\{ j
    \in \left\{ 1 .. s \right\} | w_j \neq 0 \right\}$.

For any input $\textbf{x}$ for which $c$ gives positive output there must be at
least one split node $c_i$ with positive output, while the output of all split
nodes for any other input must be less than or equal to zero. More formally, we
define the set of positive inputs as $\mathcal{X}^+ = \left \{ \mathbf{x} \in
    \{-1,1\}^s | f_{\mathbf{w}}(\mathbf{x}) > 0 \right \}$ and the set of negative
inputs as $\mathcal{X}^- = \mathcal{X} \setminus \mathcal{X}^+$, and define the
following two constraints on our collection of split nodes:
\begin{equation}\label{eq:pos-conj-split-constraints}
    \forall \mathbf{x}^+ \in \mathcal{X}^+. \exists i. f_{\tilde{\mathbf{w}}_i}(\mathbf{x}^+) > 0
    \qquad \text{and} \qquad
    \forall \mathbf{x}^- \in \mathcal{X}^-.  \forall i.
    f_{\tilde{\mathbf{w}}_i}(\mathbf{x}^-) \leq 0
\end{equation}

A \emph{split weight set} $\mathcal{W} = \{ \tilde{\mathbf{w}}_1,
    \tilde{\mathbf{w}}_2, \ldots \}$ for $c$ is the collection of weight tensors
associated with the split nodes $\{c_1, c_2, \ldots\}$. A valid
$\mathcal{W}$ can be constructed by associating each positive example
$\mathbf{x}_{i} \in \mathcal{X}^{+}$ to a unique split node $c_i$ with
weight tensor $\tilde{\mathbf{w}}_{i}$, where:
\begin{equation} \label{eq:sign_match_rule}
    \forall j \in \{1 \ldots s\}. \tilde{w}_{i, j} =
    x_{i, j} \cdot \mathbbm{1} \left( x_{i, j} = \sign(w_j) \right) \cdot 6 =
    \begin{cases}
        6 \cdot x_{i, j} & \text{if } x_{i, j} = \sign(w_j) \\
        0                & \text{otherwise}
    \end{cases}
\end{equation}
This split weight set $\mathcal{W}$ constructed under
\equationref{eq:sign_match_rule} has the cardinality of $|\mathcal{X}^+|$, and
the proof that it satisfies the constraints in
\equationref{eq:pos-conj-split-constraints} is in
\appendixref{apd:pos-conj-disentangle-proof}: the positive coverage is
straightforward and we prove by contradiction for the negative coverage.
$\mathcal{W}$ can be translated to a set of rules $\mathfrak{L}$, in the form
of:
\begin{equation} \label{eq:frak-l-definition}
    \mathfrak{L} = \bigg\{
    \text{c}                              \leftarrow
    \bigwedge_{j \in \mathcal{J}_{\ell}^{+}} \text{a}_j,
    \bigwedge_{j \in \mathcal{J}_{\ell}^{-}} \aspnot\ \text{a}_j \ \biggr\rvert
    \ \ell \in \left\{1..|\mathcal{X}^+|\right\}
    \bigg\}
\end{equation}
where $\mathcal{J}_{\ell}^{+} = \{ j \in \mathcal{J} | \tilde{w}_{\ell, j} = 6 \}, \mathcal{J}_{\ell}^{-} = \{ j \in \mathcal{J} | \tilde{w}_{\ell, j} = -6 \}$.

\subsection{Optimisation} \label{sec:optimisation}
In order to reduce the size of the split weight set, we can identify and remove
redundant weight tensors. A weight tensor is redundant if its corresponding rule
is logically subsumed by the rule of another weight tensor in the same split
weight set. In this way, we can define a subsumption relation between two weight
tensors in the split weight set as:

\begin{definition} \label{def:weight-tensor-subsumption}
    A weight tensor $\tilde{\mathbf{w}}_{m}$ \textbf{subsumes}
    $\tilde{\mathbf{w}}_{n}$ if:
    \begin{equation*}
        \left( \forall j \in \mathcal{J}.
        \tilde{w}_{m, j} \in \{0, \tilde{w}_{n, j} \} \right) \wedge
        \left( 0 <
        \left| \left\{ j \in \mathcal{J} : \tilde{w}_{m, j} \neq 0 \right\} \right| <
        \left| \left\{ j \in \mathcal{J} : \tilde{w}_{n, j} \neq 0 \right\} \right|
        \right)
    \end{equation*}

\end{definition}

If $\tilde{\mathbf{w}}_{n}$, associated to $\mathbf{x}_{n} \in \mathcal{X}^+$,
is subsumed by $\tilde{\mathbf{w}}_{m}$, $\tilde{\mathbf{w}}_{m}$ must also
cover $\mathbf{x}_{n}$ (proof in \appendixref{apd:subsumption-proof}). This
allows us to reduce $\mathcal{W}$ to a smaller yet equally valid split weight
set $\mathcal{W}^*$ where no pair of weight tensors are subsumed by each other.

To compute $\mathcal{W}^{*}$, we introduce a three-step optimisation
process:

\begin{enumerate}
    \item \textbf{Compute the set of candidate indices of all small-magnitude
              weights in the original node that might not be essential to cover
              any positive examples.}\; \; The intuition is that weights with
          large absolute values contribute more to the output bivalent
          interpretation, and thus every large positive weight should be in
          all rules' positive sets of atoms $\mathcal{J}^+_{\ell}$
          (\equationref{eq:frak-l-definition}) and every negative weight
          with large absolute value should be in all rules' negative sets of
          atoms $\mathcal{J}^-_{\ell}$. A weight $w_j$ is considered as
          having a large magnitude if $|w_j| \geq \max |\mathbf{w}| / 2$,
          meaning $\bar{\mathcal{J}} = \{j \in \mathcal{J} \ | \ |w_j| <
              \max |\mathbf{w}| / 2\}$ contains all the small-magnitude weights
          that might not be essential to cover any positive examples. The
          formal proof of this step, including how the threshold value $\max
              |\mathbf{w}| / 2$ is computed, is available in
          \appendixref{apd:optimisation-proof}.

    \item \textbf{Identify optimal combinations of candidate indices that can be
              excluded.}\; \; We define an exclusion set $\mathcal{E}_{\ell}$ of
              a weight tensor $\tilde{\mathbf{w}}_{\ell}$ as the set of indices
              of the non-zero elements in the original weight tensor
              $\mathbf{w}$ which were excluded by the disentanglement process
              when constructing $\tilde{\mathbf{w}}_{\ell}$, that is, $\{ j \in
              \mathcal{J} | \tilde{w}_{\ell, j} = 0 \land w_j \neq 0 \}$. By
              Remark~\ref{remark:max-abs-w-inequality}, we can reformulate the
              definition: a set $\mathcal{E}$ is an exclusion set if and only if
              $\sum_{j \in \mathcal{E}} |w_j| < \max |\mathbf{w}| / 2$. Since
              $\max |\mathbf{w}| / 2$ is also the threshold value used to define
              $\bar{\mathcal{J}}$, an exclusion set $\mathcal{E}$ must be a
              subset of $\bar{\mathcal{J}}$, although a subset of
              $\bar{\mathcal{J}}$ is not necessarily an exclusion set. If
              $\tilde{\mathbf{w}}_{\ell}$ subsumes $\tilde{\mathbf{w}}_{\ell'}$,
              every zero element in $\tilde{\mathbf{w}}_{\ell'}$ must also be
              zero in $\tilde{\mathbf{w}}_{\ell}$, and
              $\tilde{\mathbf{w}}_{\ell}$ must have more zero elements than
              $\tilde{\mathbf{w}}_{\ell'}$, thus $\mathcal{E}_{\ell} \supset
              \mathcal{E}_{\ell'}$. We can therefore transform the original
              optimisation problem over finding $\mathcal{W}^*$ to finding $E^*
              = \{ \mathcal{E}^*_1, \ldots, \mathcal{E}^*_\ell, \ldots,
              \mathcal{E}^*_L\}$ where no pairs of elements are subset of each
              other. To find $E^*$, we can perform a search by considering
              possible candidate $\mathcal{\hat{J}} \subset \bar{\mathcal{J}}$
              and verifying it is an exclusion set using $\sum_{j \in
              \mathcal{\hat{J}}} |w_j| < \max |\mathbf{w}| / 2$. Full details of
              this search are in
              Algorithm~\ref{algo:split-conj-with-optimisation-full}.

    \item \textbf{Convert exclusion sets to weight tensors.} \; \; For each
          exclusion set $\mathcal{E}^*_\ell \in E^*$, we create a new weight
          tensor $\tilde{\mathbf{w}}_{\ell}$ where its $j$-th element is:
          \begin{equation*}
              \tilde{w}_{\ell, j} =
              \begin{cases}
                  6 \cdot \sign(w_j) & \text{if } j \in \mathcal{J} \setminus \mathcal{E}_\ell \\
                  0                  & \text{otherwise}
              \end{cases}
          \end{equation*}
\end{enumerate}

Thus, we compute $\mathcal{W}^{*} = \{ \tilde{\mathbf{w}}_{1}, \ldots,
    \tilde{\mathbf{w}}_{L} \}, L \leq |\mathcal{X}^+|$, where no pair of weight
tensors are subsumed by each other.

\section{Predicate Invention}

\begin{figure}[htbp]
    \centering
    \subfigure[No predicate invention]{
        \includegraphics[width=0.3\textwidth]{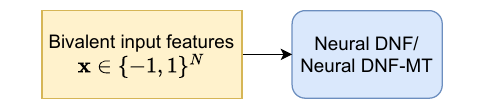}\label{fig:no-pi}
    } \hfill
    \subfigure[Neural-network-based predicate invention]{
        \includegraphics[width=0.5\textwidth]{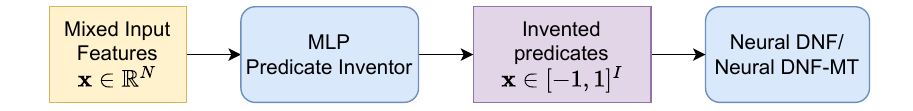}\label{fig:mlp-pi}
    } \hfill

    \subfigure[Threshold-learning predicate invention]{
        \includegraphics[width=0.8\textwidth]{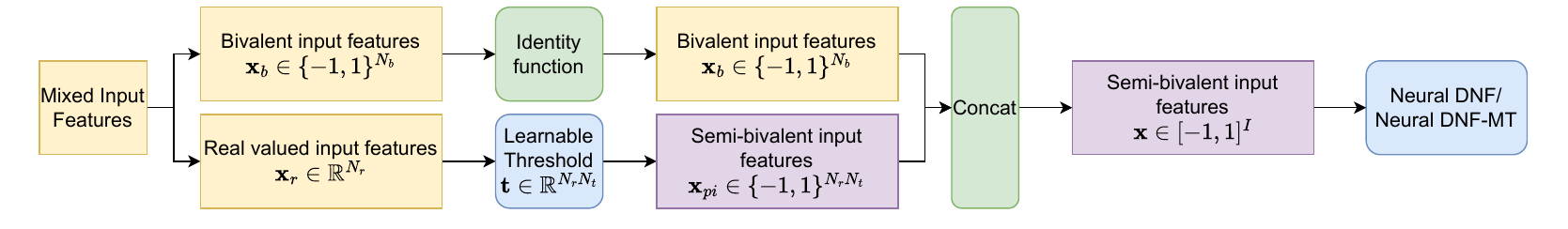}\label{fig:threshold-pi}
    }
    \caption{Architectures for various settings, where blue rounded boxes have
    learnable parameters. Neural DNF is used in binary or multilabel tasks,
    while neural DNF-MT in multiclass tasks. \ref{fig:no-pi} is the basic setup
    for bivalent inputs. \ref{fig:mlp-pi} and \ref{fig:threshold-pi} show
    how predicate invention is performed with MLPs and threshold learning.}
    \vspace{-1.5em}
\end{figure}

Neural DNF-based models require predicate invention to handle non-bivalent
inputs. \citet{pix2rule} and \citet{ndnf-mt} use neural networks
(\figureref{fig:mlp-pi}) for this purpose, extracting higher-level features as
predicates from low-level unstructured inputs. However, these neural predicates
are not inherently interpretable, and the attempt of \citet{ndnf-mt} to improve
the interpretability through an ASP optimisation process does not guarantee a
solution.

We propose an inherently interpretable approach called
\textbf{threshold-learning predicate invention} (\figureref{fig:threshold-pi}).
For each real-valued input feature $x_i$ we learn $m$ threshold values $\{t_{i,
1},\ldots t_{i,m}\}$, which are each used to create a predicate $p_{i, j} =
\tanh((x_i - t_{i, j}) / T)$ where $T \in [0.1, 1]$ is a temperature parameter.
The invented predicate $p_{i,j}$ can be interpreted as a bivalent predicate
representing the inequality $x_i > t_{i,j}$ if $p_{i,j} > 0$. We schedule the
temperature $T$ to decrease throughout training to push $p_{i,j}$ away from $0$,
thus amplifying the degree of the belief of the invented predicate. Our
threshold-learning predicate invention does not need post-training methods to be
interpretable and guarantees a solution.

\section{Evaluation}

\subsection{Experiments}

\noindent \textbf{Set up. }We train neural DNF-based models across
\emph{binary}, \emph{multiclass} and \emph{multilabel} classification tasks,
including tasks with real-valued features that require \emph{predicate
    inventions}:
\begin{itemize}
    \itemsep0em
    \item Binary: BCC\textsuperscript{$\dagger$} \citep{bcc}, Monk \citep{monk},
          Mushroom \citep{mushroom} and CDC\textsuperscript{$\dagger$}
          \citep{cdc}.
    \item Multiclass: Car \citep{car} and Covertype\textsuperscript{$\ddagger$}
          \citep{covertype}.
    \item Multilabel: ARA \citep{bn-ara}, Budding \citep{bn-budding}, Fission
          \citep{bn-fission}, and MAM \citep{bn-mam}.\footnote{These are Boolean
              Network \citep{boolean-network} datasets with ground-truth logic
              programs. Boolean Network models the state transitions of genes from
              time $t$ to $t+1$, and we treat the learning of such a dynamic system
              as a multilabel classification problem, where the labels are the gene
              states at time $t+1$.}
\end{itemize}
Datasets with $\dagger$ and $\ddagger$ require predicate inventions, and we use
threshold-learning predicate invention for $\dagger$ and MLP predicate invention
for $\ddagger$. All experiments are repeated at least 5 times. Further
experiment details can be found in \appendixref{apd:experiment-settings}.

\noindent \textbf{Results.} The first rows in
\tableref{tab:f1-scores-binary-multilabel} and \ref{tab:f1-scores-multiclass}
show that neural DNF-based models achieve competitive F1 scores compared to
standard ML approaches that also do not rely on task-specific background
knowledge, with neural DNF-based models performing the best in 7 of the 10
datasets. The neural DNF-based models match the level of expressiveness while
enabling end-to-end training with predicate invention for real-valued inputs.

\begin{table}[htbp]
    \vspace{-1em}
    \centering
    \floatconts
    {tab:f1-scores-binary-multilabel}%
    {\caption{F1 scores (mean $\pm$ ste) for binary and multilabel
            classification tasks.}}%
    {
        \resizebox{\textwidth}{!}{%
            \begin{tabular}{lcccccccccc}
                \toprule
                                    &  & \multicolumn{4}{c}{\textbf{Binary Classification}} &                                & \multicolumn{4}{c}{\textbf{Multilabel Classification}}                                                                                                                     \\ \cmidrule{3-6} \cmidrule{8-11}
                Model               &  & Monk                                               & BCC\textsuperscript{$\dagger$} & Mushroom                                               & CDC\textsuperscript{$\dagger$} &  & ARA               & Budding           & Fission           & MAM               \\ \cmidrule{1-1} \cmidrule{3-6} \cmidrule{8-11}
                NDNF                &  & 1.000 $\pm$ 0.000                                  & 0.718 $\pm$ 0.072              & 1.000 $\pm$ 0.000                                      & 0.882 $\pm$ 0.001              &  & 1.000 $\pm$ 0.000 & 1.000 $\pm$ 0.000 & 1.000 $\pm$ 0.000 & 1.000 $\pm$ 0.000 \\
                NDNF: ASP           &  & 0.992 $\pm$ 0.008                                  & 0.765 $\pm$ 0.033              & 0.971 $\pm$ 0.016                                      & 0.861 $\pm$ 0.003              &  & 0.993 $\pm$ 0.002 & 0.993 $\pm$ 0.003 & 0.987 $\pm$ 0.003 & 0.994 $\pm$ 0.003 \\ \cmidrule{1-1} \cmidrule{3-6} \cmidrule{8-11}
                MLP                 &  & 1.000 $\pm$ 0.000                                  & 0.725 $\pm$ 0.025              & 0.999 $\pm$ 0.000                                      & 0.896 $\pm$ 0.000              &  & 1.000 $\pm$ 0.000 & 1.000 $\pm$ 0.000 & 1.000 $\pm$ 0.000 & 1.000 $\pm$ 0.000 \\
                Logistic Regression &  & 0.707 $\pm$ 0.006                                  & 0.796 $\pm$ 0.020              & 1.000 $\pm$ 0.000                                      & 0.873 $\pm$ 0.000              &  & NA                & NA                & NA                & NA                \\
                SVM                 &  & 1.000 $\pm$ 0.000                                  & 0.762 $\pm$ 0.029              & 1.000 $\pm$ 0.000                                      & 0.894 $\pm$ 0.000              &  & NA                & NA                & NA                & NA                \\
                Random Forest       &  & 0.984 $\pm$ 0.005                                  & 0.753 $\pm$ 0.012              & 1.000 $\pm$ 0.000                                      & 0.889 $\pm$ 0.000              &  & 1.000 $\pm$ 0.000 & 0.996 $\pm$ 0.000 & 0.992 $\pm$ 0.001 & 0.998 $\pm$ 0.001 \\
                Decision Tree       &  & 0.877 $\pm$ 0.027                                  & 0.766 $\pm$ 0.030              & 1.000 $\pm$ 0.000                                      & 0.871 $\pm$ 0.002              &  & 1.000 $\pm$ 0.000 & 0.972 $\pm$ 0.001 & 0.954 $\pm$ 0.002 & 0.950 $\pm$ 0.002 \\
                OSDT                &  & 1.000 $\pm$ 0.000                                  & NA                             & 0.893 $\pm$ 0.000                                      & NA                             &  & NA                & NA                & NA                & NA                \\
                \bottomrule
            \end{tabular}
        }
    }
\end{table}

The second rows of Table \ref{tab:f1-scores-binary-multilabel} and
\ref{tab:f1-scores-multiclass} present the performance of the novel optimised
logical interpretation of our models as logic programs (NDNF: ASP and NDNF: LI
resp.), which in most cases leads to a minor performance drop compared to the
neural DNF-based models but at the the benefit of interpretability.
\tableref{tab:thresholding-disentanglement-comparison} shows that our new
interpretation method has a much smaller reduction in performance (right column)
compared with the pre-existing threshold method (left column) in all but one
dataset (Covertype). This exception may relate to the lack of faithfulness
guarantee for conjunctive node disentanglement when using predicate invention,
along with a higher task complexity.

\begin{table}[htbp]
    \centering
    \begin{minipage}[t]{0.48\textwidth}
        \caption{F1 scores (mean $\pm$ ste) for multiclass tasks. `NDNF-MT: LI'
            is the logical interpretation of neural DNF-MT models.} {\scriptsize
            \label{tab:f1-scores-multiclass}
            \begin{tabular}{lcc}
                \toprule
                Model         & Car               & Covertype\textsuperscript{$\ddagger$} \\
                \midrule
                NDNF-MT       & 0.963 $\pm$ 0.001 & 0.859 $\pm$ 0.003                     \\
                NDNF-MT: LI   & 0.935 $\pm$ 0.005 & 0.823 $\pm$ 0.003                     \\ \midrule
                MLP           & 0.869 $\pm$ 0.005 & 0.828 $\pm$ 0.001                     \\
                Log. Reg.     & 0.891 $\pm$ 0.003 & 0.758 $\pm$ 0.000                     \\
                SVM           & 0.937 $\pm$ 0.004 & 0.716 $\pm$ 0.001                     \\
                Random Forest & 0.954 $\pm$ 0.002 & 0.942 $\pm$ 0.000                     \\
                Decision Tree & 0.959 $\pm$ 0.004 & 0.914 $\pm$ 0.002                     \\
                \bottomrule
            \end{tabular}
        }
    \end{minipage}\hfill%
    \begin{minipage}[t]{0.48\textwidth}
        \caption{F1 score drops after thresholding ($f1_{t} - f1_{thresh}$) and
            disentanglement ($f1_{t} - f1_{disent}$) discretisation.} {\scriptsize
            \label{tab:thresholding-disentanglement-comparison}
            \resizebox{\textwidth}{!}{%
                \begin{tabular}{llcc}
                    \toprule
                                                                                              & Dataset                           & $f1_{train} - f1_{thresh}$ & $f1_{train} - f1_{disent}$ \\ \midrule
                    \parbox[t]{2mm}{\multirow{4}{*}{\rotatebox[origin=c]{90}{\textbf{Bin.}}}} & Monk                              & $0.076 \pm 0.027$          & \bm{$0.008 \pm 0.008$}     \\
                                                                                              & BCC\textsuperscript{$\dagger$}    & $0.050 \pm 0.038$          & \bm{$0.003 \pm 0.022$}     \\
                                                                                              & Mush                              & $0.316 \pm 0.099$          & \bm{$0.029 \pm 0.016$}     \\
                                                                                              & CDC\textsuperscript{$\dagger$}    & $0.031 \pm 0.007$          & \bm{$0.021 \pm 0.003$}     \\ \midrule
                    \parbox[t]{2mm}{\multirow{2}{*}{\rotatebox[origin=c]{90}{\textbf{MC.}}}}  & Car                               & $0.122 \pm 0.012$          & \bm{$0.027 \pm 0.005$}     \\
                                                                                              & Cover\textsuperscript{$\ddagger$} & \bm{$0.036 \pm 0.003$}     & $0.043 \pm 0.004$          \\ \midrule
                    \parbox[t]{2mm}{\multirow{4}{*}{\rotatebox[origin=c]{90}{\textbf{ML.}}}}  & ARA                               & $0.085 \pm 0.009$          & \bm{$0.007 \pm 0.002$}     \\
                                                                                              & Budd                              & $0.159 \pm 0.004$          & \bm{$0.007 \pm 0.003$}     \\
                                                                                              & Fission                           & $0.141 \pm 0.007$          & \bm{$0.013 \pm 0.003$}     \\
                                                                                              & MAM                               & $0.045 \pm 0.004$          & \bm{$0.005 \pm 0.003$}     \\
                    \bottomrule
                \end{tabular}
            }
        }
    \end{minipage}
\end{table}

To assess the compactness of the logical interpretations induced by our
disentangled neural DNF-based models we compare the length of the rules of its
ASP program (full ASP programs for binary/multilabel tasks, ASP programs for
conjunctive layers in multiclass task) with the tree sizes of decision trees and
OSDTs trained on the same datasets\footnote{We focus only on decision trees and
    OSDTs as direct competitors of neural DNF-based models, as they also provide
    rule-based decision processes, whereas other used ML baselines lack such
    explicit interpretability. More discussion on model interpretability can be
    found in \appendixref{apd:model-interpretability}.}. For a decision tree's
decision path, each node in the path can be seen as a condition clause of an if
statement, thus making the decision tree's depth an equivalent metric to ASP
rule length. The number of branches indicates the number of different decision
paths, making it equivalent to the number of rules in an ASP program.
\tableref{tab:rule-length-comparison} shows our models' interpretations are
consistently more compact than those of decision trees, as shown by them having
fewer rules and both the average and maximum rule length being lower. Comparing
to OSDT's, we have comparable compactness (we are more compact for the Monk
dataset while OSDT is more compact for the Mushroom dataset) whilst achieving
superior performance.

\vspace{-1em}
\begin{table}[htbp]
    \centering
    \floatconts
    {tab:rule-length-comparison} {\caption{Comparing compactness of neural
            DNF-based models' logic interpretations vs. decision trees', i.e.
            max. rule length/tree depth, avg. rule length/tree depth, and no.
            rules/branches. The cell marked with * refers to OSDT's
            performance.}%
    }
    {\tiny
        \vspace{-2em}
        \resizebox{\textwidth}{!}{%
            \begin{tabular}{llcccccccc}
                \toprule
                                                                                          & \multirow{3}{*}{Dataset}        & \multicolumn{2}{c}{Max. Rule Length/Max Depth} &                    & \multicolumn{2}{c}{Avg. Rule Length/Avg Depth} &                                    & \multicolumn{2}{c}{No. Rules/No. Branches}                                                                \\
                                                                                          &                                 & NDNF/                                          & DT/                &                                                & NDNF/                              & DT/                                        &  & NDNF/                              & DT/                  \\
                                                                                          &                                 & NDNF-MT: ASP                                   & *OSDT              &                                                & NDNF-MT: ASP                       & *OSDT                                      &  & NDNF-MT: ASP                       & *OSDT                \\ \midrule
                \parbox[t]{2mm}{\multirow{6}{*}{\rotatebox[origin=c]{90}{\textbf{Bin.}}}} & Monk                            & \multirow{2}{*}{$2.778 \pm 0.148$}             & $10.000 \pm 0.283$ &                                                & \multirow{2}{*}{$2.042 \pm 0.053$} & $7.842 \pm 0.098$                          &  & \multirow{2}{*}{$4.000 \pm 0.000$} & $49.000 \pm 7.071$   \\
                                                                                          &                                 &                                                & *$4.000 \pm 0.000$ &                                                &                                    & *$3.286 \pm 0.000$                         &  &                                    & *$7.000 \pm 0.000$   \\ \cmidrule{3-4} \cmidrule{6-7} \cmidrule{9-10}
                                                                                          & BCC\textsuperscript{$\dagger$}  & $3.200 \pm 0.179$                              & $6.600 \pm 0.219$  &                                                & $2.267 \pm 0.121$                  & $4.833 \pm 0.062$                          &  & $5.000 \pm 0.566$                  & $18.200 \pm 0.657$   \\ \cmidrule{3-4} \cmidrule{6-7} \cmidrule{9-10}
                                                                                          & Mush                            & \multirow{2}{*}{$4.250 \pm 0.225$}             & $7.400 \pm 0.358$  &                                                & \multirow{2}{*}{$2.982 \pm 0.122$} & $4.907 \pm 0.132$                          &  & \multirow{2}{*}{$3.750 \pm 0.242$} & $18.000 \pm 1.095$   \\
                                                                                          &                                 &                                                & *$2.400 \pm 0.219$ &                                                &                                    & *$1.900 \pm 0.128$                         &  &                                    & *$3.400 \pm 0.219$   \\ \cmidrule{3-4} \cmidrule{6-7} \cmidrule{9-10}
                                                                                          & CDC\textsuperscript{$\dagger$}  & $4.800 \pm 0.276$                              & $40.600 \pm 0.607$ &                                                & $2.021 \pm 0.076$                  & $21.804 \pm 0.021$                         &  & $27.600 \pm 0.851$                 & $31394 \pm 257.045$  \\ \midrule
                \parbox[c]{2mm}{\multirow{2}{*}{\rotatebox[origin=c]{90}{\textbf{MC.}}}}  & Car                             & $8.500 \pm 0.153$                              & $13.000 \pm 0.283$ &                                                & $4.432 \pm 0.075$                  & $9.462 \pm 0.060$                          &  & $25.688 \pm 0.943$                 & $109.200 \pm 1.927$  \\
                                                                                          & Cov\textsuperscript{$\ddagger$} & $4.300 \pm 0.202$                              & $25.200 \pm 0.657$ &                                                & $1.778 \pm 0.041$                  & $14.541 \pm 0.047$                         &  & $38.900 \pm 1.090$                 & $868.400 \pm 8.083$  \\ \midrule
                \parbox[t]{2mm}{\multirow{4}{*}{\rotatebox[origin=c]{90}{\textbf{ML.}}}}  & ARA                             & $3.900 \pm 0.095$                              & $13.000 \pm 0.000$ &                                                & $2.050 \pm 0.024$                  & $9.724 \pm 0.006$                          &  & $26.700 \pm 0.318$                 & $581.800 \pm 2.125$  \\
                                                                                          & Bud                             & $4.000 \pm 0.000$                              & $12.000 \pm 0.000$ &                                                & $3.126 \pm 0.036$                  & $10.937 \pm 0.007$                         &  & $48.400 \pm 1.070$                 & $1521.300 \pm 6.406$ \\
                                                                                          & Fis                             & $6.000 \pm 0.000$                              & $10.000 \pm 0.000$ &                                                & $2.852 \pm 0.020$                  & $9.308 \pm 0.018$                          &  & $21.800 \pm 0.276$                 & $392.500 \pm 4.401$  \\
                                                                                          & MAM                             & $4.000 \pm 0.000$                              & $10.000 \pm 0.000$ &                                                & $2.645 \pm 0.014$                  & $0.897 \pm 0.018$                          &  & $21.500 \pm 0.212$                 & $424.200 \pm 5.261$  \\ \bottomrule
            \end{tabular}
        }
    }
\end{table}

To summarise, the neural DNF-based models demonstrate comparable performance to
other methods that also do not rely on background knowledge and train well
together with end-to-end predicate invention, while also providing compact
logical representations at a minimal performance cost. In particular, our new
disentanglement approach reduces such performance loss when discretising the
neural DNF-based models compared to the previous thresholding method.


\subsection{Discussion}

\textbf{Limitations when Disentangling Disjunctive Layers.} \; \; We observe
entanglements involving both conjunctive and disjunctive nodes lead to imperfect
disentanglement: \exampleref{ex:nested-entanglement} shows a case where
disjunctive nodes assign weights to utilise small-magnitude activations of
conjunctive nodes to match the target. As discussed in
\sectionref{sec:disentangle-conj}, the disentanglement requires node inputs to
be exactly $\{-1, 1\}^N$. Despite using an auxiliary loss to encourage $\pm 1$
conjunctive activations, this constraint is not guaranteed for the disjunctive
nodes.  This prevents applying the same disentanglement method to the
disjunctive layer, and we find that splitting disjunctive nodes performs worse
than simply thresholding in our experiments. Future works can investigate how to
avoid these intermediate-valued cases.

\begin{figure}[h]
    \centering
    \includegraphics[width=0.6\textwidth]{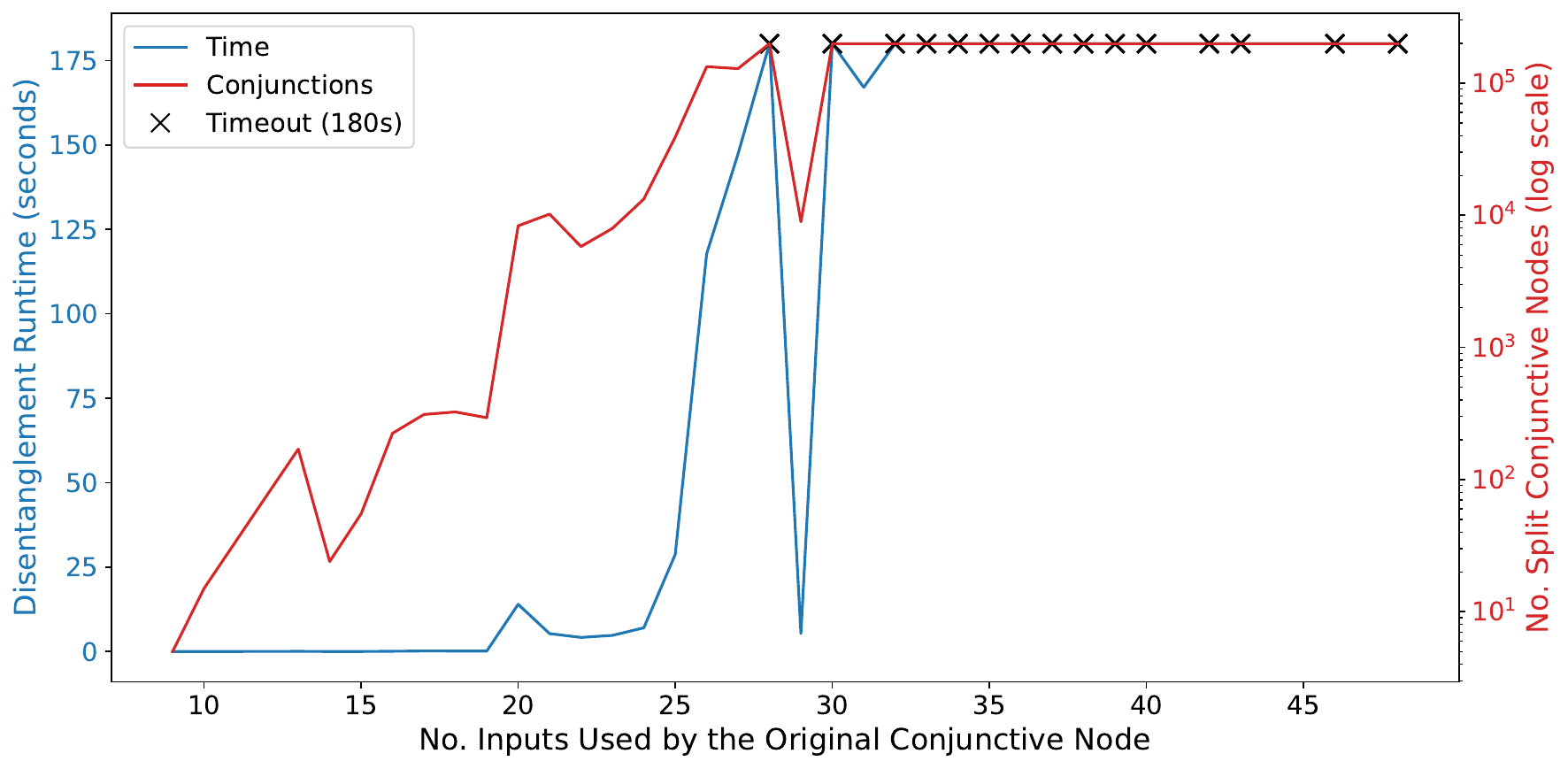}
    \caption{Time taken (blue) to disentangle the conjunctive nodes with
        different numbers of inputs used, and the number of split conjunctions
        (red) after disentanglement, in a trained neural DNF model. A max
        runtime of 180s is set for disentanglement.}
    \label{fig:conjunction-disentanglement-runtime}
    \vspace{-3em}
\end{figure}

\noindent \textbf{Scalability of Disentanglement.} \; \; The disentanglement
method's runtime grows exponentially with the number of inputs used in a node,
as shown in \figureref{fig:conjunction-disentanglement-runtime}, since it may
check all possible weight combinations that trigger the node to fire for the
worst-case scenario. This makes it less scalable than the thresholding
discretisation method, which scales linearly with weight magnitudes. However,
this limitation only affects model interpretation, not learning capability, and
can be mitigated through model sparsification.

\begin{example-box}{Nested Entanglement}{nested-entanglement}

Consider two conjunctive nodes with weights $\textbf{w}_{c1} = [-1.33, 0, 0,
    1.01, -1.44]$ and $\textbf{w}_{c2} = [-2.02, -0.79, -0.79, 0.71, -1.52]$
respectively; and two disjunctive nodes connected to those conjunctive
nodes, with weights $\textbf{w}_{d1} = [3.43, 1.28]$ and $\textbf{w}_{d2} =
    [0, 3.56]$ respectively. There is a nested entanglement among all the nodes
here: $c1$ subsumes $c2$, and $d1$ uses a combination of $c1$ and $c2$ to
match the target label. The disjunctive nodes' bivalent interpretations
change depending on whether the input activations from the conjunctive nodes
are soft-valued or strictly $\pm 1$, even when the interpretations of the
conjunctive nodes' activations are the same (differences marked in red in
the table below).

\begin{table}[H]
    \centering
    \scriptsize
    \begin{tabular}{ccccccccc}
        \toprule
        a1                        & a2          & a3        & a4           & a5                       & c1           & c2           & d1                       & d2                       \\ \hline
        -1                        & -1          & -1        & -1           & -1                       & -0.52/$\bot$ & 0.54/$\top$  & {\color{red}0.17/$\top$} & 1.00/$\top$              \\
        \multicolumn{5}{c}{-}     & -1/$\bot$   & 1/$\top$  &
        {\color{red}-0.70/$\bot$} & 1.00/$\bot$
        \\ \hline
        -1                        & -1          & -1        & 1            & 1                        & -0.89/$\bot$ & -0.77/$\bot$ & -0.99/$\bot$             & {\color{red}0.67/$\top$} \\
        \multicolumn{5}{c}{-}     & -1/$\bot$   & -1/$\bot$ & -1.00/$\bot$ & {\color{red}0.00/$\bot$}                                                                                     \\
        \bottomrule
    \end{tabular}
\end{table}

\end{example-box}

\section{Conclusion}

We propose a disentanglement discretisation method that enables more faithful
translation of neural DNF-based models into bivalent logic representations.
Additionally, we introduce an interpretable-by-design predicate invention method
that learns threshold values end-to-end, effectively converting real-valued
features into inequalities as boolean predicates. Our experiments across various
real-world classification tasks demonstrate that the disentanglement method
provides compact and interpretable logical representations for the neural
DNF-based models without requiring task-specific background knowledge, while
improving performance over the existing thresholding method. We suggest that
further research into training regimes may help optimise the trade-off between
performance and interpretability for neural DNF-based models.

\acks{This work is supported in part by DEVCOM Army Research Lab under
W911NF2220243; EPSRC projects EP/X040518/1 and EP/Y037421/1; JST CREST Grant
Number JPMJCR22D3; JSPS KAKENHI Grant Number JP25K03190; and the NII Internship
Program. We acknowledge also the project NIHR i4i: Artificial Intelligence to
support cancer early diagnosis in general practice (NIHR207533). The views
expressed in this paper are those of the author(s) and not necessarily those of
the NIHR or the Department of Health and Social Care. }


\bibliography{nesy2025}

\begin{thebibliography}{32}
\providecommand{\natexlab}[1]{#1}
\providecommand{\url}[1]{\texttt{#1}}
\expandafter\ifx\csname urlstyle\endcsname\relax
  \providecommand{\doi}[1]{doi: #1}\else
  \providecommand{\doi}{doi: \begingroup \urlstyle{rm}\Url}\fi

\bibitem[Baugh et~al.(2023)Baugh, Cingillioglu, and Russo]{ns-classifications}
Kexin~Gu Baugh, Nuri Cingillioglu, and Alessandra Russo.
\newblock Neuro-symbolic rule learning in real-world classification tasks.
\newblock In Andreas Martin, Hans-Georg Fill, Aurona Gerber, Knut Hinkelmann, Doug Lenat, Reinhard Stolle, and Frank van Harmelen, editors, \emph{Proceedings of the AAAI 2023 Spring Symposium on Challenges Requiring the Combination of Machine Learning and Knowledge Engineering (AAAI-MAKE 2023)}, volume Vol-3433. CEUR Workshop Proceedings, 2023.
\newblock URL \url{https://ceur-ws.org/Vol-3433/paper12.pdf}.

\bibitem[Baugh et~al.(2025)Baugh, Dickens, and Russo]{ndnf-mt}
Kexin~Gu Baugh, Luke Dickens, and Alessandra Russo.
\newblock Neural dnf-mt: A neuro-symbolic approach for learning interpretable and editable policies.
\newblock In \emph{Proceedings of the 24th International Conference on Autonomous Agents and Multiagent Systems}, AAMAS '25, page 252–260, Richland, SC, 2025. International Foundation for Autonomous Agents and Multiagent Systems.
\newblock ISBN 9798400714269.

\bibitem[Blackard(1998)]{covertype}
Jock Blackard.
\newblock {Covertype}.
\newblock UCI Machine Learning Repository, 1998.
\newblock {DOI}: https://doi.org/10.24432/C50K5N.

\bibitem[Bohanec(1988)]{car}
Marko Bohanec.
\newblock {Car Evaluation}.
\newblock UCI Machine Learning Repository, 1988.
\newblock {DOI}: https://doi.org/10.24432/C5JP48.

\bibitem[Chaos et~al.(2006)Chaos, Aldana, Espinosa-Soto, de~Le{\'o}n, Arroyo, and Alvarez-Buylla]{bn-ara}
{\'A}lvaro Chaos, Max Aldana, Carlos Espinosa-Soto, Berenice Garc{\'i}a~Ponce de~Le{\'o}n, Adriana~Garay Arroyo, and Elena~R. Alvarez-Buylla.
\newblock From genes to flower patterns and evolution: Dynamic models of gene regulatory networks.
\newblock \emph{Journal of Plant Growth Regulation}, 25\penalty0 (4):\penalty0 278--289, Dec 2006.
\newblock ISSN 1435-8107.
\newblock \doi{10.1007/s00344-006-0068-8}.
\newblock URL \url{https://doi.org/10.1007/s00344-006-0068-8}.

\bibitem[Cingillioglu and Russo(2021)]{pix2rule}
Nuri Cingillioglu and Alessandra Russo.
\newblock {pix2rule: End-to-end Neuro-symbolic Rule Learning}.
\newblock In Artur d'Avila Garcez and Ernesto Jiménez-Ruiz, editors, \emph{Proceedings of the 15th International Workshop on Neural-Symbolic Learning and Reasoning (NeSy 2021) as part of the 1st International Joint Conference on Learning \& Reasoning (IJCLR 2021)}. CEUR Workshop Proceedings, 2021.
\newblock URL \url{https://ceur-ws.org/Vol-2986/paper3.pdf}.

\bibitem[Cunnington et~al.(2023{\natexlab{a}})Cunnington, Law, Lobo, and Russo]{ffnsl}
Daniel Cunnington, Mark Law, Jorge Lobo, and Alessandra Russo.
\newblock Ffnsl: Feed-forward neural-symbolic learner.
\newblock \emph{Machine Learning}, 112\penalty0 (2):\penalty0 515--569, Feb 2023{\natexlab{a}}.
\newblock ISSN 1573-0565.
\newblock \doi{10.1007/s10994-022-06278-6}.
\newblock URL \url{https://doi.org/10.1007/s10994-022-06278-6}.

\bibitem[Cunnington et~al.(2023{\natexlab{b}})Cunnington, Law, Lobo, and Russo]{ns-asp-raw-data}
Daniel Cunnington, Mark Law, Jorge Lobo, and Alessandra Russo.
\newblock Neuro-symbolic learning of answer set programs from raw data.
\newblock In Edith Elkind, editor, \emph{Proceedings of the Thirty-Second International Joint Conference on Artificial Intelligence, {IJCAI-23}}, pages 3586--3596. International Joint Conferences on Artificial Intelligence Organization, 8 2023{\natexlab{b}}.
\newblock \doi{10.24963/ijcai.2023/399}.
\newblock URL \url{https://doi.org/10.24963/ijcai.2023/399}.
\newblock Main Track.

\bibitem[Davidich and Bornholdt(2008)]{bn-fission}
Maria~I. Davidich and Stefan Bornholdt.
\newblock Boolean network model predicts cell cycle sequence of fission yeast.
\newblock \emph{PLOS ONE}, 3\penalty0 (2):\penalty0 1--8, 02 2008.
\newblock \doi{10.1371/journal.pone.0001672}.
\newblock URL \url{https://doi.org/10.1371/journal.pone.0001672}.

\bibitem[De~Raedt et~al.(2007)De~Raedt, Kimmig, and Toivonen]{problog}
Luc De~Raedt, Angelika Kimmig, and Hannu Toivonen.
\newblock Problog: a probabilistic prolog and its application in link discovery.
\newblock In \emph{Proceedings of the 20th International Joint Conference on Artifical Intelligence}, IJCAI'07, page 2468–2473, San Francisco, CA, USA, 2007. Morgan Kaufmann Publishers Inc.

\bibitem[Delfosse et~al.(2023)Delfosse, Shindo, Dhami, and Kersting]{nudge}
Quentin Delfosse, Hikaru Shindo, Devendra Dhami, and Kristian Kersting.
\newblock Interpretable and explainable logical policies via neurally guided symbolic abstraction.
\newblock In A.~Oh, T.~Naumann, A.~Globerson, K.~Saenko, M.~Hardt, and S.~Levine, editors, \emph{Advances in Neural Information Processing Systems}, volume~36, pages 50838--50858. Curran Associates, Inc., 2023.
\newblock URL \url{https://proceedings.neurips.cc/paper_files/paper/2023/file/9f42f06a54ce3b709ad78d34c73e4363-Paper-Conference.pdf}.

\bibitem[Evans and Grefenstette(2018)]{delta-ilp}
Richard Evans and Edward Grefenstette.
\newblock Learning explanatory rules from noisy data.
\newblock \emph{Journal of Artificial Intelligence Research}, 61:\penalty0 1--64, 2018.

\bibitem[Fauré et~al.(2006)Fauré, Naldi, Chaouiya, and Thieffry]{bn-mam}
Adrien Fauré, Aurélien Naldi, Claudine Chaouiya, and Denis Thieffry.
\newblock Dynamical analysis of a generic boolean model for the control of the mammalian cell cycle.
\newblock \emph{Bioinformatics}, 22\penalty0 (14):\penalty0 e124--e131, 07 2006.
\newblock ISSN 1367-4803.
\newblock \doi{10.1093/bioinformatics/btl210}.
\newblock URL \url{https://doi.org/10.1093/bioinformatics/btl210}.

\bibitem[Gao et~al.(2022)Gao, Wang, Cao, and Inoue]{dlfit}
Kun Gao, Hanpin Wang, Yongzhi Cao, and Katsumi Inoue.
\newblock Learning from interpretation transition using differentiable logic programming semantics.
\newblock \emph{Machine Learning}, 111\penalty0 (1):\penalty0 123--145, Jan 2022.
\newblock ISSN 1573-0565.
\newblock \doi{10.1007/s10994-021-06058-8}.
\newblock URL \url{https://doi.org/10.1007/s10994-021-06058-8}.

\bibitem[Glanois et~al.(2022)Glanois, Jiang, Feng, Weng, Zimmer, Li, Liu, and Hao]{hri}
Claire Glanois, Zhaohui Jiang, Xuening Feng, Paul Weng, Matthieu Zimmer, Dong Li, Wulong Liu, and Jianye Hao.
\newblock Neuro-symbolic hierarchical rule induction.
\newblock In Kamalika Chaudhuri, Stefanie Jegelka, Le~Song, Csaba Szepesvari, Gang Niu, and Sivan Sabato, editors, \emph{Proceedings of the 39th International Conference on Machine Learning}, volume 162 of \emph{Proceedings of Machine Learning Research}, pages 7583--7615. PMLR, 17--23 Jul 2022.
\newblock URL \url{https://proceedings.mlr.press/v162/glanois22a.html}.

\bibitem[Guide(1981)]{mushroom}
Audobon Society~Field Guide.
\newblock {Mushroom}.
\newblock UCI Machine Learning Repository, 1981.
\newblock {DOI}: https://doi.org/10.24432/C5959T.

\bibitem[Hu et~al.(2019)Hu, Rudin, and Seltzer]{osdt}
Xiyang Hu, Cynthia Rudin, and Margo Seltzer.
\newblock Optimal sparse decision trees.
\newblock In H.~Wallach, H.~Larochelle, A.~Beygelzimer, F.~d\textquotesingle Alch\'{e}-Buc, E.~Fox, and R.~Garnett, editors, \emph{Advances in Neural Information Processing Systems}, volume~32, pages 7265--7273. Curran Associates, Inc., 2019.
\newblock URL \url{https://proceedings.neurips.cc/paper_files/paper/2019/file/ac52c626afc10d4075708ac4c778ddfc-Paper.pdf}.

\bibitem[Kauffman(1969)]{boolean-network}
Stuart Kauffman.
\newblock Homeostasis and differentiation in random genetic control networks.
\newblock \emph{Nature}, 224\penalty0 (5215):\penalty0 177--178, Oct 1969.
\newblock ISSN 1476-4687.
\newblock \doi{10.1038/224177a0}.
\newblock URL \url{https://doi.org/10.1038/224177a0}.

\bibitem[Kimura et~al.(2021)Kimura, Ono, Chaudhury, Kohita, Wachi, Agravante, Tatsubori, Munawar, and Gray]{nsrl-fol}
Daiki Kimura, Masaki Ono, Subhajit Chaudhury, Ryosuke Kohita, Akifumi Wachi, Don~Joven Agravante, Michiaki Tatsubori, Asim Munawar, and Alexander Gray.
\newblock Neuro-symbolic reinforcement learning with first-order logic.
\newblock In Marie-Francine Moens, Xuanjing Huang, Lucia Specia, and Scott Wen-tau Yih, editors, \emph{Proceedings of the 2021 Conference on Empirical Methods in Natural Language Processing}, pages 3505--3511, Online and Punta Cana, Dominican Republic, 2021. Association for Computational Linguistics.
\newblock \doi{10.18653/v1/2021.emnlp-main.283}.
\newblock URL \url{https://aclanthology.org/2021.emnlp-main.283}.

\bibitem[Law et~al.(2018)Law, Russo, and Broda]{ilasp}
Mark Law, Alessandra Russo, and Krysia Broda.
\newblock The complexity and generality of learning answer set programs.
\newblock \emph{Artificial Intelligence}, 259:\penalty0 110--146, 2018.
\newblock ISSN 0004-3702.
\newblock \doi{https://doi.org/10.1016/j.artint.2018.03.005}.
\newblock URL \url{https://www.sciencedirect.com/science/article/pii/S000437021830105X}.

\bibitem[Li et~al.(2004)Li, Long, Lu, Ouyang, and Tang]{bn-budding}
Fangting Li, Tao Long, Ying Lu, Qi~Ouyang, and Chao Tang.
\newblock The yeast cell-cycle network is robustly designed.
\newblock \emph{Proceedings of the National Academy of Sciences}, 101\penalty0 (14):\penalty0 4781--4786, 2004.
\newblock \doi{10.1073/pnas.0305937101}.
\newblock URL \url{https://www.pnas.org/doi/abs/10.1073/pnas.0305937101}.

\bibitem[Lifschitz(2019)]{asp}
Vladimir Lifschitz.
\newblock \emph{Answer set programming}.
\newblock Springer Nature, Cham, Switzerland, 2019.
\newblock \doi{https://doi.org/10.1007/978-3-030-24658-7}.

\bibitem[Muggleton and {de Raedt}(1994)]{ilp}
Stephen Muggleton and Luc {de Raedt}.
\newblock Inductive logic programming: Theory and methods.
\newblock \emph{The Journal of Logic Programming}, 19-20:\penalty0 629--679, 1994.
\newblock ISSN 0743-1066.
\newblock \doi{https://doi.org/10.1016/0743-1066(94)90035-3}.
\newblock URL \url{https://www.sciencedirect.com/science/article/pii/0743106694900353}.
\newblock Special Issue: Ten Years of Logic Programming.

\bibitem[Patr{\'i}cio et~al.(2018)Patr{\'i}cio, Pereira, Cris{\'o}stomo, Matafome, Gomes, Sei{\c{c}}a, and Caramelo]{bcc}
Miguel Patr{\'i}cio, Jos{\'e} Pereira, Joana Cris{\'o}stomo, Paulo Matafome, Manuel Gomes, Raquel Sei{\c{c}}a, and Francisco Caramelo.
\newblock Using resistin, glucose, age and bmi to predict the presence of breast cancer.
\newblock \emph{BMC Cancer}, 18\penalty0 (1):\penalty0 29, Jan 2018.
\newblock ISSN 1471-2407.
\newblock \doi{10.1186/s12885-017-3877-1}.
\newblock URL \url{https://doi.org/10.1186/s12885-017-3877-1}.

\bibitem[Phua and Inoue(2024)]{delta-lfit-2}
Yin~Jun Phua and Katsumi Inoue.
\newblock Variable assignment invariant neural networks for learning logic programs.
\newblock In Tarek~R. Besold, Artur d'Avila Garcez, Ernesto Jimenez-Ruiz, Roberto Confalonieri, Pranava Madhyastha, and Benedikt Wagner, editors, \emph{Neural-Symbolic Learning and Reasoning}, pages 47--61, Cham, 2024. Springer Nature Switzerland.
\newblock ISBN 978-3-031-71167-1.

\bibitem[Rudin(2019)]{interpretable-model}
Cynthia Rudin.
\newblock Stop explaining black box machine learning models for high stakes decisions and use interpretable models instead.
\newblock \emph{Nature Machine Intelligence}, 1\penalty0 (5):\penalty0 206--215, May 2019.
\newblock ISSN 2522-5839.
\newblock \doi{10.1038/s42256-019-0048-x}.
\newblock URL \url{https://doi.org/10.1038/s42256-019-0048-x}.

\bibitem[Sen et~al.(2022)Sen, de~Carvalho, Riegel, and Gray]{lnn-ilp}
Prithviraj Sen, Breno~WSR de~Carvalho, Ryan Riegel, and Alexander Gray.
\newblock Neuro-symbolic inductive logic programming with logical neural networks.
\newblock In \emph{Proceedings of the AAAI Conference on Artificial Intelligence}, volume~36, pages 8212--8219, 2022.

\bibitem[Sha et~al.(2023)Sha, Shindo, Kersting, and Dhami]{nesy-pi}
Jingyuan Sha, Hikaru Shindo, Kristian Kersting, and Devendra~Singh Dhami.
\newblock Neural-symbolic predicate invention: Learning relational concepts from visual scenes.
\newblock In Artur d'Avila Garcez and Ernesto Jiménez-Ruiz, editors, \emph{NeSy 2023, 17th International Workshop on Neural-Symbolic Learning and Reasoning}, pages 103--117. CEUR Workshop Proceedings, 2023.
\newblock URL \url{https://ceur-ws.org/Vol-3432/paper8.pdf}.

\bibitem[Shindo et~al.(2021)Shindo, Nishino, and Yamamoto]{dilp-structured}
Hikaru Shindo, Masaaki Nishino, and Akihiro Yamamoto.
\newblock Differentiable inductive logic programming for structured examples.
\newblock In \emph{Proceedings of the AAAI Conference on Artificial Intelligence}, volume~35, pages 5034--5041, 2021.

\bibitem[Shindo et~al.(2023)Shindo, Pfanschilling, Dhami, and Kersting]{alpha-ilp}
Hikaru Shindo, Viktor Pfanschilling, Devendra~Singh Dhami, and Kristian Kersting.
\newblock Alpha ilp: thinking visual scenes as differentiable logic programs.
\newblock \emph{Machine Learning}, 112\penalty0 (5):\penalty0 1465--1497, May 2023.
\newblock ISSN 1573-0565.
\newblock \doi{10.1007/s10994-023-06320-1}.
\newblock URL \url{https://doi.org/10.1007/s10994-023-06320-1}.

\bibitem[Teboul(2021)]{cdc}
Alex Teboul.
\newblock Diabetes health indicators dataset.
\newblock Kaggle, 2021.
\newblock URL \url{https://www.kaggle.com/datasets/alexteboul/diabetes-health-indicators-dataset}.

\bibitem[Wnek(1993)]{monk}
J.~Wnek.
\newblock {MONK's Problems}.
\newblock UCI Machine Learning Repository, 1993.
\newblock {DOI}: https://doi.org/10.24432/C5R30R.

\end{thebibliography}

\newpage

\appendix

\section{Semi-symbolic Node and Neural DNF-based Model} \label{apd:additional-background}

A neural Disjunctive Normal Form model \cite{pix2rule} is a neural model with
nodes that can behave as a semi-symbolic conjunction or disjunction. The
formalisation of a semi-symbolic node is listed in \equationref{eq:ss-node}, and
the neural DNF model itself is constructed with a layer of conjunctive nodes
followed by a layer of disjunctive nodes. The inputs to the semi-symbolic node
are strictly bounded in $[-1, 1]$, where the extreme value $1$ ($-1$) is
interpreted as $\top$ ($\bot$) and the intermediate values represent weaker
strengths of belief. The node activation $\hat{y}$ can be interpreted similarly
but cannot take the extreme values $\pm 1$. A bivalent interpretation of
$\hat{y}$ is to treat $\hat{y} > 0$ ($\le 0$) as $\top$ ($\bot$). $\delta$
induces behaviour similar to a conjunction (disjunction) when $\delta = 1$ ($=
-1$). During training, the absolute values of $\delta$ in both layers gradually
increase to 1 (controlled by a scheduler) for better learning. \citet{pix2rule}
shows how a neural DNF model can be used in binary classification.
\citet{ns-classifications} proposes an extended model called the neural DNF-EO
for multi-class classifications. It has a non-trainable constraint layer after
the disjunctive layer to ensure a logically mutually exclusive output.
\citet{ndnf-mt} proposes a neural DNF-MT model. The $\tanh$ activation in the
final disjunctive layer is replaced with a mutex-tanh activation function, so
that the model can represent mutually-exclusive probability distributions as its
outputs for reinforcement learning. \citet{ndnf-mt} also shows how end-to-end
predicate-invention can be performed with the model. All these methods interpret
trained neural DNF models with `deterministic' behaviour as logical rules
represented in Answer Set Programming \citep{asp}. Neural DNF-MT model also
supports probabilistic interpretation in ProbLog \citep{problog} for its
disjunctive layer but its conjunctive layer is still translated to deterministic
rules.

\begin{example-box}{Bivalent logic translation}{bivalent-logic-translation}

    A symbolisation function $\sigma$ maps names of inputs ($x_i$) or nodes'
    bivalent interpretations ($b_{ci}$/$b_{di}$ for conj./disj. nodes) to symbolic
    predicate names: e.g. $\sigma(x_i) = \text{a}_i$, $\sigma(b_{ci}) = \text{c}_i$
    and $\sigma(b_{di}) = \text{d}_i$. Under this $\sigma$, a conjunctive
    semi-symbolic node with weights $[6, 0, 0, -6]$ is translated into $\text{c}
    \leftarrow \text{a}_1, \aspnot\ \text{a}_4.$ Similarly, a disjunctive
    semi-symbolic node with conjunctive nodes as inputs and weights $ [0, -6, 6]$ is
    translated into $\{ \text{d} \leftarrow \aspnot\ \text{c}_2.\ \text{d}
    \leftarrow \text{c}_3. \}$.

\end{example-box}

\section{Disentanglement of Positively Connected Conjunction}

This section follows the setting in \sectionref{sec:disentangle-conj}, where we
want to disentangle a conjunctive node positively connected to a disjunctive
node, i.e. the weight connecting them is positive.

Recall the goal split weight set $\mathcal{W} = \{ \tilde{\mathbf{w}}_1,
    \tilde{\mathbf{w}}_2, \ldots, \tilde{\mathbf{w}}_L \}$ in
\sectionref{sec:smaller-split}. The weights must have the form:
\begin{equation} \label{eq:def-new-weights}
    \forall \ell \in \{1..L\}, i \in \{1..s\}. \begin{cases}
        \tilde{w}_{\ell, i} \in \{-6, 0, 6\} & \text{if } i \in \mathcal{J} \\
        \tilde{w}_{\ell, i} = 0              & \text{otherwise}
    \end{cases}.
\end{equation}
Constraints \ref{eq:pos-conj-split-constraints} can be rewritten as:
\begin{align}
    \forall \mathbf{x} \in \mathcal{X}^+.
    \exists      & \ell \left [
        \sum_{j \in \mathcal{J}} \tilde{w}_{\ell, j} x_j + \left(6 - 6\sum_{j \in \mathcal{J}} |\tilde{w}_{\ell, j}|\right) = 6
    \right ] \label{eq:w_positive} \\
    \forall \mathbf{x} \in \mathcal{X}^-.
    \neg \exists & \ell \left [
        \sum_{j \in \mathcal{J}} \tilde{w}_{\ell, j} x_j + \left(6 - 6\sum_{j \in \mathcal{J}} |\tilde{w}_{\ell, j}|\right) > -6
        \right ]\label{eq:w_negative}
\end{align}
This means that for every positive example ($\mathbf{x}^+$) at least one rule is
activated, and for every negative example ($\mathbf{x}^-$) none of the rules are
activated.

\subsection{Proof of Coverage of Split} \label{apd:pos-conj-disentangle-proof}

Recall the splitting method in \sectionref{sec:smaller-split}. Here we prove
that the set $\mathcal{W} = \{\tilde{\mathbf{w}}_1, \ldots,
    \tilde{\mathbf{w}}_{| \mathcal{X}^+ |}\}$ constructed under
\equationref{eq:sign_match_rule} satisfies Conditions~(\ref{eq:w_positive}) and
(\ref{eq:w_negative}), i.e. all positive examples are covered and no negative
example is covered.

\begin{proof} \textbf{Positive Coverage}

    The positive coverage is straightforward. For any positive example
    $\mathbf{x}_i \in \mathcal{X}^+$, take $\tilde{\mathbf{w}}_i$ associated to
    $\mathbf{x}_i$. By \equationref{eq:sign_match_rule}, we have:
    \begin{equation}
        \begin{aligned}
              & \sum_{j \in \mathcal{J}} \tilde{w}_{i,j} x_{i,j} + \left( 6 - \sum_{j \in \mathcal{J}} |\tilde{w}_{i,j}| \right)                                                             \\
            = & \ 6 \sum_{j \in \mathcal{J}} \mathbbm{1} \left( x_{i,j} = \sign(w_j) \right) + \left( 6 - 6 \sum_{j \in \mathcal{J}} \mathbbm{1} \left( x_{i,j} = \sign(w_j) \right) \right) \\
            = & \ 6
        \end{aligned}
    \end{equation}
    The set $\{\tilde{\mathbf{w}}_1, \ldots, \tilde{\mathbf{w}}_{| \mathcal{X}^+
            |}\}$ will cover all positive examples $\mathbf{x}_i \in \mathcal{X}^+$.
\end{proof}

\begin{proof} \textbf{Negative Coverage}

    We will prove that the set $\{\tilde{\mathbf{w}}_1, \ldots,
        \tilde{\mathbf{w}}_{| \mathcal{X}^+ |}\}$ constructed under
    \equationref{eq:sign_match_rule} does not cover any of the negative examples
    by contradiction.

    Assume that $\mathbf{x}_m \in \mathcal{X}^-$ is covered by
    $\tilde{\mathbf{w}}_i$, which is associated to positive example
    $\mathbf{x}_i \in \mathcal{X}^+$. Then the raw output of
    $\tilde{\mathbf{w}}_i$ over $\mathbf{x}_m$ should be 6 to cover
    $\mathbf{x}_m$:
    \begin{align}
                    & f_{\mathbf{w}_i}(\mathbf{x}_m) = 6 \nonumber                                                                                     \\
                    & \sum_{j \in \mathcal{J}} \tilde{w}_{i, j} x_{m, j} + \left(6 - \sum_{j \in \mathcal{J}} |\tilde{w}_{i, j}| \right) = 6 \nonumber \\
        \Rightarrow & \sum_{j \in \mathcal{J}} \tilde{w}_{i, j} x_{m, j} = \sum_{j \in \mathcal{J}} |\tilde{w}_{i, j}| \label{eq:wixi=abs_wi}
    \end{align}

    Substitute the definition of $\tilde{w}_{i, j}$ in
    \equationref{eq:sign_match_rule} into \equationref{eq:wixi=abs_wi}, and we
    have:
    \begin{equation} \label{eq:xin_xi}
        \sum_{j \in \mathcal{J}} x_{i,j} \cdot \mathbbm{1}
        \left( x_{i,j} = \sign (w_j) \right)
        \cdot x_{m,j} = \sum_{j \in \mathcal{J}} \mathbbm{1} \left( x_{i,j} = \sign (w_j) \right)
    \end{equation}

    When $x_{i,j} \neq \sign (w_j)$, $\mathbbm{1} \left( x_{i,j} = \sign (w_j)
        \right) = 0$, and will not contribute to the summation term in the
    L.H.S of \equationref{eq:xin_xi}. Thus, for \equationref{eq:xin_xi}
    to hold, this has to hold:
    \begin{equation} \label{eq:xin_xi_relation}
        \forall j \in \mathcal{J}.\ \left[
            x_{i, j} = \sign (w_j) \implies
            x_{m, j} = x_{i, j} = \sign (w_j)
            \right]
    \end{equation}
    From (\ref{eq:xin_xi_relation}) we know that $\{j \in \mathcal{J} | x_{i, j}
        = \sign (w_j) \} \subset \{j \in \mathcal{J} | x_{m, j} = \sign (w_j)
        \}$. So $\{j \in \mathcal{J} | x_{i, j} \neq \sign (w_j) \} \supset \{j
        \in \mathcal{J} | x_{m, j} \neq \sign (w_j) \}$. The subset cannot be
    equal because $\mathbf{x}_i \neq \mathbf{x}_m$. And we can split the set
    $\{j \in \mathcal{J} | x_{m, j} = \sign (w_j) \}$ as the following:
    \begin{equation*}
        \begin{aligned}
            \{j \in \mathcal{J} | x_{m, j} = \sign (w_j) \} & \\
            = \left\{j \in \mathcal{J} | x_{m, j} = \sign (w_j) \wedge x_{i, j} = \sign (w_j) \right\}
                                                            &
            \cup \left\{j \in \mathcal{J} | x_{m, j} = \sign (w_j) \wedge x_{i, j} \neq \sign (w_j)  \right\}
        \end{aligned}
    \end{equation*}
    \figureref{fig:proof-pos-split} helps to demonstrate the relations between
    the sets.

    \begin{figure}[h]
        \centering
        \includegraphics[width=\textwidth]{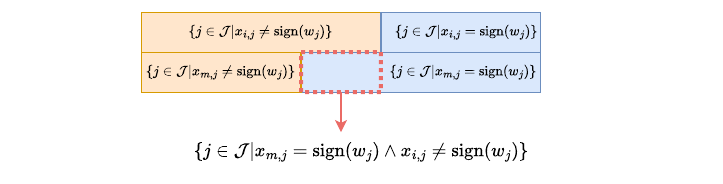}
        \label{fig:proof-pos-split}
    \end{figure}

    Now, we compute the raw output of the original conjunctive node with weight
    $\mathbf{w}$ over negative example $\mathbf{x}_m$:
    \begin{equation}
        \begin{aligned}
            f_{\mathbf{w}}(\mathbf{x}_m) = & \sum_{j \in \mathcal{J}} w_j x_{m,j} + \beta                                                                                            \\
            =                              & \sum_{j \in \mathcal{J}: x_j = \sign(w_j)} |w_j| - \sum_{j \in \mathcal{J}: x_j \neq \sign(w_j)} |w_j| + \beta  \label{eq:x_activation} \\
        \end{aligned}
    \end{equation}

    We can rewrite \equationref{eq:x_activation} as the following:
    \begin{equation} \label{eq:x_activation_long}
        f_{\mathbf{w}}(\mathbf{x}_m) = \sum_{
            \substack{j \in \mathcal{J}: \\
                x_{m, j} = \sign(w_j), \\
                x_{i, j} = \sign (w_j)}
        } |w_j| \  + \sum_{
            \substack{j \in \mathcal{J}: \\
                x_{m, j} = \sign(w_j), \\
                x_{i, j} \neq \sign (w_j)}
        } |w_j| \ -  \sum_{
            \substack{j \in \mathcal{J}: \\
                x_{m, j} \neq \sign(w_j), \\
                x_{i, j} \neq \sign (w_j)}
        } |w_j| \ + \beta
    \end{equation}
    If we flip the sign of the second term on the R.H.S. of
    \equationref{eq:x_activation_long}, since it is guaranteed to be greater
    than 0, the following inequality holds:
    \begin{equation} \label{eq:ax_inequality}
        f_{\mathbf{w}}(\mathbf{x}) > \sum_{
            \substack{j \in \mathcal{J}: \\
                x_{m, j} = \sign(w_j), \\
                x_{i, j} = \sign (w_j)}
        } |w_j| \  - \sum_{
            \substack{j \in \mathcal{J}: \\
                x_{m, j} = \sign(w_j), \\
                x_{i, j} \neq \sign (w_j)}
        } |w_j| \ -  \sum_{
            \substack{j \in \mathcal{J}: \\
                x_{m, j} \neq \sign(w_j),\\
                x_{i, j} \neq \sign (w_j)}
        } |w_j| \ + \beta
    \end{equation}
    Using the illustration in \figureref{fig:proof-pos-split}, we can transform
    the R.H.S of (\ref{eq:ax_inequality}):
    \begin{equation}
        \begin{aligned}
            \sum_{j \in \mathcal{J}: x_{i, j} = \sign (w_j) } |w_j| \ -
            \sum_{j \in \mathcal{J}: x_{i, j} \neq \sign (w_j)} |w_j| \ + \beta
            = f_{\mathbf{w}}(\mathbf{x}_i) > 0.
        \end{aligned}
    \end{equation}

    Now we have $f_{\mathbf{w}}(\mathbf{x}_m) > f_{\mathbf{w}}(\mathbf{x}_i) >
        0$, but this contradicts our assumption that $\mathbf{x}_m \in
        \mathcal{X}^-$ such that $f_{\mathbf{w}}(\mathbf{x}_m) \leq 0$. So we
    just proved that no negative examples will be covered by the set
    $\{\tilde{\mathbf{w}}_1, \ldots, \tilde{\mathbf{w}}_{| \mathcal{X}^+
            |}\}$ constructed under \equationref{eq:sign_match_rule}.

\end{proof}

We have shown that the split weight set $\{\tilde{\mathbf{w}}_1, \ldots,
    \tilde{\mathbf{w}}_{| \mathcal{X}^+ |}\}$ constructed under
\equationref{eq:sign_match_rule} will cover all positive examples and none
of the negative examples.

\subsection{Optimisation} \label{apd:optimisation-proof}

\subsubsection{Subsumption} \label{apd:subsumption-proof}

Following the setting in \sectionref{sec:optimisation}, if
$\tilde{\mathbf{w}}_{p}$ associated to $\mathbf{x}_{p}$ is subsumed by
$\tilde{\mathbf{w}}_{q}$, $\tilde{\mathbf{w}}_{q}$ also covers the positive
example $\mathbf{x}_{p}$.

\begin{proof}
    We need to show that the raw output of $\tilde{\mathbf{w}}_{q}$ with input
    $\mathbf{x}_{p}$ is 6 to prove its coverage.
    \begin{equation}
        \begin{aligned}
              & \sum_{j \in \mathcal{J}} \tilde{w}_{q, j} x_{p, j} + \left (6 - \sum_{j \in \mathcal{J}} |\tilde{w}_{q, j}| \right )                                                                                                                  \\
            = & \ 6 \sum_{j \in \mathcal{J}} \mathbbm{1} \left (  \tilde{w}_{q, j} = \tilde{w}_{p, j} \right ) \tilde{w}_{p, j} x_{p, j} +                                                                                                            \\
              & \left( 6 - 6 \sum_{j \in \mathcal{J}}   \mathbbm{1} \left (  \tilde{w}_{q, j} = \tilde{w}_{p, j} \right ) | \tilde{w}_{p, j} |  \right)                              &  & \scriptstyle(\text{by def. of subsumption})                 \\
            = & \ 6 \sum_{j \in \mathcal{J}} \mathbbm{1} \left (  \tilde{w}_{q, j} = \tilde{w}_{p, j} \right )  \mathbbm{1} \left(x_{p, j} = \sign(w_j)\right) +                                                                                      \\
              & \left( 6 - 6 \sum_{j \in \mathcal{J}}   \mathbbm{1} \left (  \tilde{w}_{q, j} = \tilde{w}_{p, j} \right ) \mathbbm{1} \left (x_{p, j} = \sign(w_j) \right )  \right) &  & \scriptstyle(\text{by Equation~(\ref{eq:sign_match_rule})}) \\
            = & \ 6
        \end{aligned}
    \end{equation}
\end{proof}

\subsubsection{Remarks for Computing Smaller Split Weight Set} \label{apd:optimisation-remarks}

\begin{remark} \label{remark:max-abs-w-inequality}
    Given a sample $\mathbf{x}^{(k)} \in \{-1, 1\}^{s}$, we can split
    $\mathcal{J}$ into two sets, based on whether the signs of $w_j$ and
    $x_j^{(k)}$ are matching:
    \begin{align} \label{eq:split-D}
        \mathcal{J}^{(k)+} & = \left\{
        j \in \mathcal{J}  \; \middle\vert \; w_j x^{(k)}_j > 0 \right\}  \nonumber \\
        \mathcal{J}^{(k)-} & = \left\{
        j \in \mathcal{J}  \; \middle\vert \; w_j x^{(k)}_j < 0 \right\}            \\
        \mathcal{J}^{(k)+} & \cup \mathcal{J}^{(k)-} = \mathcal{J} \nonumber
    \end{align}

    We can rewrite \equationref{eq:conjuncitve-node-raw-output} as:
    $f_{\mathbf{w}}\left(\mathbf{x}^{(k)}\right) = \max_{j \in \mathcal{J}}
        |w_j| - 2 \sum_{j \in \mathcal{J}^{(k)-}} |w_j|$. Given $\mathbf{x}^{(k)}$,
    the bivalent interpretation of this node $b^{(k)} = \top$ when
    $f_{\mathbf{w}}(\mathbf{x}^{(k)}) > 0$. In the context of splitting a
    positively used conjunctive node, we care about the following property:
    \begin{equation} \label{eq:max-abs-w-inequality}
        f_{\mathbf{w}}(\mathbf{x}^{(k)}) > 0 \iff \max_{j\in\mathcal{J}} |w_j| > 2 \sum_{j\in\mathcal{J}^{(k)-}} |w_j|
    \end{equation}

\end{remark}

Recall the translated rule set $\mathfrak{L}$ that covers the soft-valued truth
table of the original conjunctive node, indexed by $\ell$, defined in
\equationref{eq:frak-l-definition}. Following the splitting method described in
Section~\ref{sec:smaller-split}, at least one of the rules in $\mathfrak{L}$
will be triggered for any positive sample $\mathbf{x}^{(n)}$. If Rule $\ell$ is
triggered by $\mathbf{x}^{(n)}$, i.e.
$f_{\tilde{\mathbf{w}}_\ell}(\mathbf{x}^{(n)}) > 0$. Using
\remarkref{remark:max-abs-w-inequality}, we know that for any $j \in
    \mathcal{J}$ s.t. $w_j \geq 1/2 \max_{j\in\mathcal{J}} |w_j|$, $j \notin
    \mathcal{J}^{(n)-}$ and $j \in \mathcal{J}^{+}_{\ell}$ for all $\ell$. The
intuition is that every large positive weight should be in all rules' positive
sets of atoms $\mathcal{J}^+_{\ell}$, since the large weight indicates a heavy
contribution from the input to the output bivalent interpretation. Similarly, if
$w_j \leq -1/2 \max_{j\in\mathcal{J}} |w_j|$, then $j \in
    \mathcal{J}^{-}_{\ell}$ for all $\ell$. This property has been discussed in
\sectionref{sec:optimisation}, and we can formalise it as:

\begin{remark} \label{remark:half-max-abs-w-inclusion}%
    We split the relevant set $\mathcal{J}$ into $\mathcal{J}^+ = \{j \in
        \mathcal{J} | w_j > 0 \}$ and $\mathcal{J}^- = \{j \in \mathcal{J} | w_j
        < 0 \}$, to separate the positive and negative weights. We define the
    following sets of indices:
    \begin{align*}
        \bar{\mathcal{J}}   & = \left\{j \in \mathcal{J} \; \middle\vert \;
        |w_{j}| < \frac{1}{2} \max_{j' \in \mathcal{J}} |w_{j'}| \right\}   \\
        \bar{\mathcal{J}}^+ & = \bar{\mathcal{J}} \cap \mathcal{J}^{+}
                            & \quad \text{small but positive weights}       \\
        \bar{\mathcal{J}}^- & = \bar{\mathcal{J}} \cap \mathcal{J}^{-}
                            & \quad \text{small but negative weights}
    \end{align*}
    The indices where the weights' absolute values are greater than half of the
    maximum absolute weight should be included in all split rules:
    \begin{align*}
        \forall \ell. j. \left[ j \in (\mathcal{J}^{+}\setminus\bar{\mathcal{J}}^{+}) \implies j \in \mathcal{J}^{+}_{\ell} \right] \\
        \forall \ell. j. \left[ j \in (\mathcal{J}^{-}\setminus\bar{\mathcal{J}}^{-}) \implies j \in \mathcal{J}^{-}_{\ell} \right]
    \end{align*}
\end{remark}

\subsection{Pseudocode}

\algorithmref{algo:split-conj-with-optimisation-full} shows the full pseudocode
on how to split a positively used conjunctive node with optimisation.

\newpage

\begin{algorithm2e}[H]
    \caption{The full algorithm on splitting a positively used conjunctive node
        with optimisation}
    \label{algo:split-conj-with-optimisation-full}
    \SetAlgoNoLine
    \LinesNumbered

    \KwIn{Weight $\mathbf{w}$ of a conjunctive node positively connected
        to a disjunctive node}

    \KwOut{List of valid split weights}

    Compute $\mathcal{J} \gets \{j \in \{1..s\} | w_j \neq 0\}$

    Compute $\bar{\mathcal{J}} \gets \{j \in \mathcal{J} \ | \ |w_j| <
        \max |\mathbf{w}| / 2\}$ (\remarkref{remark:half-max-abs-w-inclusion})

    \lIf{$\bar{\mathcal{J}} = \emptyset$}{
        \Return{ $[\text{sign}(\mathbf{w}) \cdot 6]$ }
    }

    \tcp{Use breadth-first search to explore subsets of $\bar{\mathcal{J}}$ that
        valid\\ exclusion candidates $\mathcal{E}$ while maximising its
        cardinality, where:\\
        $\sum_{j \in \mathcal{E}} |w_j| < \max |\mathbf{w}| / 2$}

    $valid\_splits \gets \{\}$, $queue \gets$ empty OrderedDict \;

    Add all $j \in \bar{\mathcal{J}}$ as a set $\{j\}$ to $queue$ with parent = None \;

    \lForAll{$j \in \bar{\mathcal{J}}$}{
        $queue.add(\{j\}, \text{parent} = \text{None})$
    }

    \While{$queue$ is not empty}{
        Get the head item $(removal\_indices, parent)$ from $queue$ \;

        $new\_half\_max \gets \max | \mathbf{w} | / 2 -  \sum | \mathbf{w}[removal\_indices] |$\;

        $\mathcal{J}_{new} \gets \{ j \in \mathcal{J} \ | \ j \notin removal\_indices \text{ and } |w_j| < new\_half\_max \}$ \;

        \uIf{$new\_half\_max \leq 0$ \textbf{or} $\mathcal{J}_{new} = \emptyset$}{

            \tcp{ The current `removal\_indices' is a not valid exclusion set or\\
                there are no more elements in $\hat{\mathcal{J}}$ that
                can be added to the\\ exclusion set.}

            \lIf{$new\_half\_max > 0$}{
                $valid\_splits.\text{add}(removal\_indices)$
            }
            \lElseIf{$parent$ is not None}{
                $valid\_splits.\text{add}(parent)$
            }
            \Continue \;
        }

        \tcp{Keep exploring bigger exclusion sets}
        $\mathcal{J}_{new} \gets \mathcal{J}_{new} - removal\_indices$ \;

        \uIf{$\mathcal{J}_{new}$ is empty}{
            $valid\_splits.\text{add}(removal\_indices)$ \;

            \Continue \;
        }

        \ForEach{$j \in \mathcal{J}_{new}$}{
            $new\_removal \gets \text{sort}(removal\_indices \cup \{j\})$ \;

            \uIf{$new\_removal \notin queue$}{
                $queue.\text{add}(new\_removal, \text{parent} = removal\_indices)$ \;
            }
        }

    }

    $split\_tensors \gets []$

    \ForEach{valid exclusion set $\mathcal{E}$}{
        Create new tensor $\tilde{\mathbf{w}}$ where:

        \quad - $j \in \mathcal{E}$: $\tilde{w}_j = 0$

        \quad - $j \in \mathcal{J} \setminus \mathcal{E}$: $\tilde{w}_j = 6 \cdot \text{sign}(w_j)$

        Add $\tilde{\mathbf{w}}$ to split tensors\;
    }

    \Return{split\_tensors}\;

\end{algorithm2e}

\section{Disentanglement of Negatively Used Conjunction} \label{apd:neg-conj-disentangle}

Now we consider a pruned conjunctive node parameterised with weight $\mathbf{w}$
with a total of $s$ inputs, but it's connected to a disjunctive node with a
negative weight. This means that the disjunctive node will fire only when the
conjunction is not true.

We reuse the definition of relevant set $\mathcal{J}$ and the two sets of inputs
based on the soft-valued truth table $\mathbb{X}^+$ and $\mathbb{X}^-$. We want
a set of split weights $\{\tilde{\mathbf{w}}_1, \ldots\}$ with
$\tilde{\mathbf{w}}_i \in \{-6, 0, 6\}^d$ such that:
\begin{align}
    \forall \mathbf{x} \in \mathcal{X}^-.
    \exists      & \ell \left [
        \sum_{j\in\mathcal{J}} \tilde{w}_{\ell, j} x_j + \left(6 - \sum_{j\in\mathcal{J}}|\tilde{w}_{\ell, j}|\right) = 6
    \right ] \label{eq:neg/w_negative} \\
    \forall \mathbf{x} \in \mathcal{X}^+.
    \neg \exists & \ell \left [
        \sum_{j\in\mathcal{J}} \tilde{w}_{\ell, j} x_j + \left(6 - \sum_{j\in\mathcal{J}}|\tilde{w}_{\ell, j}|\right) > -6
        \right ] \label{eq:neg/w_positive}
\end{align}
Note that we reverse the condition compared to the positive case in
\sectionref{sec:disentangle-conj}. If an input results in a less-or-equal-to-0
output for conjunctive node with $\mathbf{w}$, the connected disjunctive node
will fire, and we want a new conjunctive node that \emph{fires} under this input
and connect it to the same disjunctive node but \emph{positively}
(\equationref{eq:neg/w_negative}). And the new conjunctive node should
\emph{not} cover any of the \emph{positive} examples
(\equationref{eq:neg/w_positive}).

Here we purpose a valid way of constructing such split weights. For each
negative example $\mathbf{x}_i \in \mathcal{X}^-$ (which we do want to cover
with the new conjunctive nodes), we associate a split weight
$\tilde{\mathbf{w}}_i$ to it, where:
\begin{equation} \label{eq:neg/sign_mismatch_rule}
    \forall j \in \{1 \ldots s\}. \tilde{w}_{i, j} =
    x_{i, j} \cdot \mathbbm{1} \left( x_{i, j} \neq \sign(w_j) \right) \cdot 6 =
    \begin{cases}
        6 \cdot x_{i, j} & \text{if } x_{i, j} \neq \sign(w_j) \\
        0                & \text{otherwise}
    \end{cases}
\end{equation}
Note that each element $\tilde{w}_i$ will still be in the set of $\{-6, 0,
    6\}$. Doing so we create a set of split weights $\mathcal{W} =
    \{\tilde{\mathbf{w}}_1, \ldots, \tilde{\mathbf{w}}_{|\mathcal{X}^-|}\}$.

\subsection{Negative Coverage: Negative Examples Fire the Split Nodes}

Take a negative example $\mathbf{x}_i \in \mathcal{X}^-$. By construction
described in \equationref{eq:neg/sign_mismatch_rule}, we have:
\begin{equation}
    \begin{aligned}
          & \sum_{j \in \mathcal{J}} \tilde{w}_{i, j} x_{i, j} + \left (6 - \sum_{j\in\mathcal{J}} |\tilde{w}_{i, j}| \right )                                                              \\
        = & \ 6 \sum_{j\in\mathcal{J}} \mathbbm{1} \left(x_{i, j} \neq \sign(w_j)\right) + \left( 6 - 6 \sum_{j\in\mathcal{J}} \mathbbm{1} \left (x_{i, j} \neq \sign(w_j) \right ) \right) \\
        = & \ 6
    \end{aligned}
\end{equation}
Thus, the negative example $\mathbf{x}_i$ will fire the new conjunction node
parameterised by $\tilde{\mathbf{w}}_i$.

Similarly, we can prove that if $\tilde{\mathbf{w}}_p$ subsumes
$\tilde{\mathbf{w}}_q$, it also fires with the input of negative example
$\mathbf{x}_q$ that is associated to $\tilde{\mathbf{w}}_q$, and
$\tilde{\mathbf{w}}_q$ becomes not necessary in the final program. So we can
compute a smaller final program: $\{ \tilde{\mathbf{w}}_i \}_{i\in
    \mathcal{I}}$, where $\mathcal{I} = \{ i \in \{ 1..|\mathcal{X}^-|\} \ | \
    \tilde{\mathbf{w}}_i$ is not subsumed by another rule$\}$.

\subsection{Positive Coverage: Positive Examples Do Not Fire Split Nodes}

Now we prove that $\mathcal{W}$ does not fire for any of the positive examples,
by contradiction.

\begin{figure}[h]
    \centering
    \includegraphics[width=\textwidth]{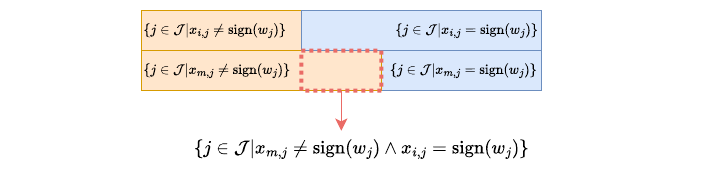}
    \label{fig:proof-neg-split}
\end{figure}

Assume that the split weight $\tilde{\mathbf{w}}_{i}$ associated to negative
example $\mathbf{x}_{i} \in \mathcal{X}^-$ is fired for $\mathbf{x}_m \in
    \mathcal{X}^+$. Then the raw output of $\tilde{\mathbf{w}}_i$ over
$\mathbf{x}_m$ should be 6 to cover $\mathbf{x}_m$:
\begin{align}
                & f_{\tilde{\mathbf{w}}}(\mathbf{x}_m) = 6                                                                                                  \nonumber \\
                & \sum_{j \in \mathcal{J}} \tilde{w}_{i, j} x_{m, j} + \left(6 - \sum_{j \in \mathcal{J}} |\tilde{w}_{i, j}| \right) = 6 \nonumber                    \\
    \Rightarrow & \sum_{j \in \mathcal{J}} \tilde{w}_{i, j} x_{m, j} = \sum_{j \in \mathcal{J}} |\tilde{w}_{i, j}| \label{eq:neg/wixi=abs_wi}
\end{align}
Substitute the definition of $\tilde{w}_{i, j}$ in
\equationref{eq:neg/sign_mismatch_rule} into \equationref{eq:neg/wixi=abs_wi},
and we have:
\begin{equation}\label{eq:neg/xiq_xi}
    \sum_{j \in \mathcal{J}} x_{i, j} \cdot \mathbbm{1} \left( x_{i, j} \neq \sign (w_j) \right) \cdot x_{m, j} = \sum_{j \in \mathcal{J}} \mathbbm{1} \left( x_{i, j} \neq \sign (w_j) \right)
\end{equation}
For \equationref{eq:neg/xiq_xi} to hold, the following has to hold:
\begin{equation} \label{eq:neg/xiq_xi_relation}
    \forall j \in \mathcal{J}.\; \left[
        x_{m, j} = x_{i, j}  \implies
        x_{i, j} \neq \sign (w_j)
        \right]
\end{equation}
From (\ref{eq:neg/xiq_xi_relation}) we know that $\{j \in \mathcal{J} | x_{i, j}
    \neq \sign (w_j) \} \subset \{j \in \mathcal{J} | x_{m, j} \neq \sign (w_j)
    \}$. So $\{j \in \mathcal{J} | x_{m, j} = \sign (w_j) \} \supset \{j \in
    \mathcal{J} | x_{i, j} = \sign (w_j) \}$, which means that $x_{m, j} =
    \sign(w_j)$ holds for any $j \in \mathcal{J}$ such that $x_{i, j} = \sign
    (w_j)$. The subset cannot be equal because $\mathbf{x}_i \neq \mathbf{x}_m$.
And we can split the set $\{j \in \mathcal{J} | x_{m, j} \neq \sign (w_j)
    \}$ like this:
\begin{equation*}
    \begin{aligned}
        \{j \in \mathcal{J} | x_{m, j} \neq \sign (w_j) \} & \\
        = \left\{j \in \mathcal{J} | x_{m, j} \neq \sign (w_j) \wedge x_{i, j} \neq \sign (w_j) \right\}
                                                           &
        \cup \left\{j \in \mathcal{J} | x_{m, j} \neq \sign (w_j) \wedge x_{i, j} \neq \sign (w_j)  \right\}
    \end{aligned}
\end{equation*}
\figureref{fig:proof-neg-split} helps to demonstrate the relations between the
    sets.

Now, we compute the raw output of the original conjunctive node with weight
$\mathbf{w}$ over positive example $\mathbf{x}_m$:
\begin{equation}
    \begin{aligned}
        f_{\mathbf{w}}(\mathbf{x}_m) = & \sum_{j \in \mathcal{J}} w_j x_{m, j} + \beta                                                                                                    \\
        =                              & \sum_{j \in \mathcal{J}: x_{m, j} = \sign(w_j)} |w_j| - \sum_{j \in \mathcal{J}: x_{m, j} \neq \sign(w_j)} |w_j| + \beta  \label{eq:neg/x_activation} \\
    \end{aligned}
\end{equation}
We can rewrite \equationref{eq:neg/x_activation} as the following:
\begin{equation} \label{eq:neg/x_activation_long}
    f_{\mathbf{w}}(\mathbf{x}_m) =
        \sum_{
            \substack{
                j \in \mathcal{J}: \\
                x_{m, j} = \sign(w_j),\\
                x_{i, j} = \sign (w_j)
            }
        } |w_j| \; -
        \sum_{
            \substack{
                j \in \mathcal{J}: \\
                x_{m, j} \neq \sign(w_j), \\
                x_{i, j} = \sign (w_j)
            }
        } |w_j| \; -
        \sum_{
            \substack{
                j \in \mathcal{J}:\\
                x_{m, j} \neq \sign(w_j),\\
                x_{i, j} \neq \sign (w_j)
            }
        } |w_j| \ + \beta
\end{equation}
If we flip the sign of the second term in
\equationref{eq:neg/x_activation_long}, which is guaranteed to be greater than
0, the following inequality holds:
\begin{equation} \label{eq:neg/ax_inequality}
    f_{\mathbf{w}}(\mathbf{x}_m) <
        \sum_{
            \substack{
                j \in \mathcal{J}: \\
                x_{m, j} = \sign(w_j),\\
                x_{i, j} = \sign (w_j)
            }
        } |w_j| \; +
        \sum_{
            \substack{
                j \in \mathcal{J}: \\
                x_{m, j} \neq \sign(w_j), \\
                x_{i, j} = \sign (w_j)
            }
        } |w_j| \; -
        \sum_{
            \substack{
                j \in \mathcal{J}:\\
                x_{m, j} \neq \sign(w_j),\\
                x_{i, j} \neq \sign (w_j)
            }
        } |w_j| \ + \beta
\end{equation}
We can see that the R.H.S of Inequality~(\ref{eq:neg/ax_inequality}) is the same
as:
\begin{equation*}
    \begin{aligned}
        \sum_{j \in \mathcal{J}: x_{i, j} = \sign (w_j) } |w_j| \ - \sum_{j \in \mathcal{J}: x_{i, j} \neq \sign (w_j)} |w_j| \ + \beta
    \end{aligned}
\end{equation*}
which is exactly $f_{\mathbf{w}}(\mathbf{x}_i)$. Because $\mathbf{x}_i \in
    \mathcal{X}^-$, $f_{\mathbf{w}}(\mathbf{x}_i) < 0$.

Now we have $f_{\mathbf{w}}(\mathbf{x}_m) < f_{\mathbf{w}}(\mathbf{x}_i) < 0$,
but this contradicts our assumption that $\mathbf{x}_m \in \mathcal{X}^+$ where
$f_{\mathbf{w}}(\mathbf{x}_m) > 0$. So we prove that no positive examples will
fire any split node constructed under \equationref{eq:neg/sign_mismatch_rule}.

We have shown that $\{\tilde{\mathbf{w}}_1, \ldots,
\tilde{\mathbf{w}}_{|\mathcal{X}^-|}\}$ constructed under
\equationref{eq:neg/sign_mismatch_rule} will fire for all negative examples as
input and not for any of the positive examples as input. And by pruning split
nodes that are subsumed by other split nodes, we can also compute an equally
valid but smaller set of split weights $\{ \tilde{\mathbf{w}}_1, \ldots,
\tilde{\mathbf{w}}_{|\mathcal{Q}|}\}$ where $\mathcal{Q} = \{ q \in \{
1..|\mathcal{X}^-|\} \ | \ \tilde{\mathbf{w}}_q$ is not subsumed by another
rule$\}$.

\hfill $\blacksquare$

\section{Experiments} \label{apd:experiments}


\subsection{Interpretability of Models} \label{apd:model-interpretability}

While many models offer varying degrees of interpretability, decision trees and
neural DNF-based models stand out for their high interpretability by providing
conditional prediction in a rule-based decision-making process: decision trees
provide decision paths that can be followed from inputs to decisions; and the
neural DNF-based models are designed such that they provide rule-based
interpretations. Logistic regression and random forest are two less
interpretable models with composition nature: logistic regression shows the
additive contribution of inputs to the log-odds of the final decision; each tree
in a random forest is rule-based but the overall decision is derived from a
weighted voting system of the trees' decisions. Support Vector Machines (SVMs)
and Multi-Layer Perceptrons (MLPs) are typically not considered interpretable due
to their complex decision boundaries and lack of transparency in decision-making
processes. Because of the different characteristics, we focus on decision trees
as direct competitors to neural DNF-based models for our interpretability study,
and ignore the other models with non-rule-based interpretations.


\subsection{Experiment Settings} \label{apd:experiment-settings}

\tableref{tab:experiment-settings} shows the experiment settings across the
datasets for MLP and neural DNF-based models. For logistic regressions, SVMs,
random forests and decision trees, we run the experiments with 5-fold cross
validation.

\begin{table}[h]
    \centering
    \floatconts
    {tab:experiment-settings}%
    {\caption{Experiment settings for MLP and neural DNF-based models. Predicate
    invention requirement is only applicable to neural DNF-based models.}}%
    {
        \begin{tabular}{lccc}
            \toprule
            Dataset   & \begin{tabular}[c]{@{}c@{}}Neural DNF-based Models:\\Requires Predicate Invention?\end{tabular} & \begin{tabular}[c]{@{}c@{}}Hold-out Test /\\ Cross Validaiton\end{tabular} & Number of Runs / Folds \\ \midrule
            \multicolumn{4}{l}{\textbf{Binary}}                                                                                                                                                       \\
            Monk      & No                                                                      & Hold-out Test                                                              & 18                     \\
            BCC       & Yes                                                                     & Cross Validation                                                           & 5                      \\
            Mushroom  & No                                                                      & Hold-out Test                                                              & 16                     \\
            CDC       & Yes                                                                     & Hold-out Test                                                              & 10                     \\ \midrule
            \multicolumn{4}{l}{\textbf{Multiclass}}                                                                                                                                                   \\
            Car       & No                                                                      & Hold-out Test                                                              & 16                     \\
            Covertype & Yes                                                                     & Hold-out Test                                                              & 10                     \\ \midrule
            \multicolumn{4}{l}{\textbf{Multilabel}}                                                                                                                                                   \\
            ARA       & No                                                                      & Cross Validation                                                           & 10                     \\
            Budding   & No                                                                      & Cross Validation                                                           & 10                     \\
            Fission   & No                                                                      & Cross Validation                                                           & 10                     \\
            MAM       & No                                                                      & Cross Validation                                                           & 10                     \\ \bottomrule
        \end{tabular}
    }
\end{table}


\subsection{Examples of Logical Interpretation} \label{apd:logical-interpretation}

\subsubsection{Binary Classification}

Below is an example of the logical interpretation of a neural DNF model in the
the dataset \textbf{Monk} \citep{monk}.

\begin{lstlisting}[label=lst:monk-li]
% t is the target class predicate.
% a_i is the input feature.
t :- a_1, a_4.   t :- a_2, a_5.
t :- a_0, a_3.
t :- not a_11, not a_12, not a_13.
\end{lstlisting}

Below is an example of the logical interpretation of a neural DNF model in the
dataset \textbf{BCC} \citep{bcc}, where threshold-learning predicate invention
method is used.

\begin{lstlisting}[label=lst:bcc-li]
% Invented predicates defined by the threshold-learning
% predicate invention method.
a_2 = feature_0 > -269.8374328613281
a_3 = feature_0 > 186.82131958007812
a_6 = feature_1 > 58.69133377075195
a_6 = feature_1 > 58.69133377075195
a_8 = feature_2 > 197.28598022460938
a_14 = feature_3 > -13.131363868713379
a_28 = feature_7 > 4.719961166381836

% Rules
% t is the target class predicate.
% a_i are invented predicates shown above.
t :- a_2, not a_3, not a_6.
t :- a_8, a_14.
t :- a_2, not a_6, a_28.
\end{lstlisting}

Below is an example of the logical interpretation of a neural DNF model in the
dataset \textbf{Mushroom} \citep{mushroom}.

\begin{lstlisting}[label=lst:mushroom-li]
% t is the target class predicate.
% a_i are input features.
t :- a_50, a_97, a_103.
t :- a_46, a_89, not a_105, not a_109.
t :- not a_21, not a_24, not a_26.
\end{lstlisting}

Below is an example of the logical interpretation of a neural DNF model in the
dataset \textbf{CDC} diabetes \citep{cdc}, where threshold-learning predicate
invention method is used.

\begin{lstlisting}[label=lst:cdc-li]
% Invented predicates defined by the threshold-learning
% predicate invention method.
a_2 = feature_0 > 74.53063201904297
a_5 = feature_0 > 58.61220932006836
a_6 = feature_0 > -33.472694396972656
a_7 = feature_0 > 102.41500854492188
a_8 = feature_1 > 22.021968841552734
a_12 = feature_1 > 16.420833587646484
a_15 = feature_1 > 19.272249221801758
a_18 = feature_2 > 45.10814666748047
a_25 = feature_3 > 51.938621520996094

% Rules
% t is the target class predicate.
% a_i with i in [0, 63] are invented predicates shown above.
% a_i with i in [63, 91] are binary input features.
t :- a_35.
t :- a_38.
t :- a_52.
t :- a_47.
t :- a_43.
t :- not a_37.
t :- a_39.
t :- a_45.
t :- a_7.
t :- a_25.
t :- a_18.
t :- a_2, a_12, not a_32, not a_41.
t :- a_2, a_15, not a_32.
t :- not a_34, a_36, not a_42, not a_59.
t :- a_8, a_36, not a_42, not a_59.
t :- a_36, not a_42, a_55.
t :- a_48.
t :- not a_41, not a_42, not a_59.
t :- not a_34, a_36, not a_51.
t :- not a_49, a_53.
t :- a_5, not a_42, not a_50, not a_51, not a_57.
t :- not a_6, not a_59.
t :- a_40, not a_51.
t :- a_36, a_53.
\end{lstlisting}

\subsubsection{Multiclass Classification}

Below is an example of the logical interpretation of a neural DNF-MT model in
the dataset \textbf{Car} \citep{car}.

\begin{lstlisting}[label=lst:car-li]
% conj_i are the output predicates of the conjunctive layer.
% a_i are the input features.
conj_0 :- not a_13, not a_14.
conj_1 :- a_2, not a_5, not a_12, not a_17, a_20.
conj_1 :- a_2, not a_12, a_18.
conj_1 :- not a_7, not a_12, a_18.
conj_1 :- a_1, not a_12, a_18.
conj_1 :- a_1, not a_5, not a_16, a_20.
conj_3 :- not a_0, not a_3, a_5, not a_8, a_14, not a_17,
    a_20.
conj_3 :- a_1, a_5, not a_9, not a_12, not a_17, a_20.
conj_6 :- not a_9, not a_10, not a_13, a_17.
conj_7 :- not a_2, not a_5, a_9, not a_14, a_16, not a_18.
conj_7 :- not a_1, not a_6, not a_11, not a_14, a_16,
    not a_18.
conj_8 :- a_2, a_5, not a_12, not a_19, not a_20.
conj_8 :- a_2, a_5, a_13, not a_19.
conj_9 :- not a_7, not a_17, not a_19.
conj_9 :- not a_3, not a_7, not a_19.
conj_9 :- not a_0, not a_3, not a_19, not a_20.
conj_12 :- not a_20.
conj_13 :- a_3, not a_5, not a_6.
conj_14 :- a_1, a_5, not a_12, not a_19, not a_20.
conj_15 :- a_1, not a_7, not a_12, not a_17, a_18.
conj_16 :- not a_0.
conj_17 :- a_0, not a_7, not a_12, not a_17, not a_19.
conj_21 :- a_1, a_6, not a_12, not a_19.
conj_22 :- not a_0, not a_3, not a_4, not a_7, not a_8,
    not a_12, not a_17, not a_19, not a_20.
conj_24 :- not a_5, not a_6, a_17, a_20.
% During inference, given an input, we first compute the
% conjunctions that are activated.
% Then we generate the ProbLog rule.
% For example:
% Input (tensor):
[-1., -1.,  1., -1., -1., -1., -1.,  1., -1.,  1., -1.,
-1., -1., -1., 1., -1.,  1., -1., -1., -1.,  1.]
% Input (translated to predicates):
[a_2, a_7, a_9, a_14, a_16, a_20]
% Conjunctions that are activated:
[conj_1, conj_16]
% ProbLog rule:
0.806::class_0 ; 0.020::class_1 ; 0.171::class_2 ;
0.003::class_3 :- conj_1, conj_16.
% Final prediction: class_0, ground truth: class_0
\end{lstlisting}

Below is an example of the logical interpretation of a neural DNF-MT model in
the dataset \textbf{Covertype} \citep{covertype}, where an MLP is used as a
predicate inventor.

\begin{lstlisting}[label=lst:covertype-li]
% conj_i are the output predicates of the conjunctive layer.
% a_i are the invented predicates from the MLP predicate
% inventor.
conj_0 :- not a_14.
conj_1 :- not a_6, a_27, not a_57.
conj_2 :- a_1.
conj_3 :- not a_46, a_52.
conj_6 :- not a_26.
conj_7 :- a_42.
conj_10 :- a_24.
conj_13 :- a_0.
conj_16 :- a_12, a_36, not a_51.
conj_17 :- not a_3.
conj_18 :- not a_50, a_52.
conj_19 :- not a_18, a_62.
conj_19 :- a_34.
conj_20 :- not a_57.
conj_21 :- a_12, a_17.
conj_22 :- a_0, not a_41.
conj_23 :- not a_49, not a_61.
conj_24 :- a_20.
conj_25 :- not a_21, not a_26.
conj_26 :- a_63.
conj_27 :- not a_51, a_56.
conj_28 :- not a_26, a_39.
conj_29 :- a_39.
conj_30 :- not a_26, a_48.
conj_32 :- not a_40.
conj_33 :- a_44.
conj_34 :- a_10.
conj_35 :- not a_44.
conj_36 :- a_33.
conj_37 :- not a_42.
conj_38 :- not a_44.
conj_40 :- a_29.
conj_41 :- not a_55.
conj_42 :- not a_10, not a_58.
conj_43 :- not a_61, a_62.
conj_44 :- not a_48.
conj_45 :- a_41, not a_50.
conj_46 :- a_31.
conj_47 :- a_43.
conj_48 :- a_12.
conj_50 :- not a_54.
conj_51 :- a_19, not a_59.
conj_52 :- not a_1, not a_2, a_9, not a_51.
conj_53 :- not a_15, not a_22, not a_29, not a_38, not a_43.
conj_54 :- a_9, not a_15, a_37.
conj_55 :- a_49.
conj_56 :- not a_49, not a_53.
conj_57 :- a_36, not a_37.

% During inference, given an input, we first compute the
% invented predicates from the MLP predicate inventor.
% Then we check which conj_i are activated.
% Finally, we generate the ProbLog rule.
% For example:
% Input:
[ 1.3166, -0.5331, -0.2809, -0.1949, -0.4017,  0.2064,  1.0031,
-0.0667, -0.2774,  1.0000, -1.0000,  1.0000,  1.0000,  1.0000,
-1.0000, -1.0000]
% Invented predicates that are true:
[a_0, a_1, a_2, a_3, a_7, a_8, a_9, a_10, a_11, a_14, a_15, a_17,
a_22, a_26, a_27, a_28, a_29, a_31, a_32, a_33, a_34, a_38, a_39,
a_41, a_43, a_44, a_47, a_48, a_49, a_51, a_53, a_54, a_58, a_59,
a_60, a_61, a_63]
% Conjunctions that are activated:
[conj_1, conj_2, conj_13, conj_19, conj_20, conj_26, conj_29,
conj_32, conj_33, conj_34, conj_36, conj_37, conj_40, conj_41,
conj_45, conj_46, conj_47, conj_55]
% ProbLog rule:
0.005::class_0 ; 0.047::class_1 ; 0.000::class_2 ;
0.000::class_3 ; 0.128::class_4 ; 0.001::class_5 ;
0.820::class_6 :- conj_1, conj_2, conj_13, conj_19, conj_20,
conj_26, conj_29, conj_32, conj_33, conj_34, conj_36, conj_37,
conj_40, conj_41, conj_45, conj_46, conj_47, conj_55.
% Final prediction: class_6, ground truth: class_6
\end{lstlisting}

\subsubsection{Multilabel Classification}

The datasets in this section are Boolean Network \citep{boolean-network}
datasets, with ground-truth logic programs. Boolean Network models the state
transitions of genes from time $t$ to $t+1$, and we treat the learning of such a
dynamic system as a multilabel classification problem: the inputs are gene
states at time $t$ and the labels are the gene states at time $t+1$.

Below is an example of the logical interpretation of a neural DNF model in the
dataset \textbf{ARA} \citep{bn-ara}.

\begin{lstlisting}[label=lst:ara-li]
% l_i are the target labels.
% a_i are inputs.
l_0 :- a_1, a_6.
l_0 :- a_0, a_4, a_13, a_14.
l_0 :- a_0, a_9, a_13, a_14.
l_1 :- a_1.
l_2 :- not a_4, not a_12.
l_3 :- not a_5.
l_4 :- a_6, not a_9.
l_4 :- a_3, not a_9.
l_4 :- not a_9, not a_12.
l_5 :- not a_6.
l_6 :- not a_12.
l_6 :- not a_5.
l_7 :- not a_12.
l_8 :- a_8, not a_9.
l_8 :- a_8, not a_14.
l_9 :- not a_7, not a_12.
l_9 :- a_6, a_8.
l_9 :- a_6, not a_11.
l_9 :- not a_4, a_6.
l_9 :- a_6, not a_10.
l_12 :- not a_4, a_5, not a_6.
l_13 :- a_6, a_9.
l_13 :- a_0, a_6.
l_14 :- a_6.
\end{lstlisting}

Below is an example of the logical interpretation of a neural DNF model in the
dataset \textbf{Budding} \citep{bn-budding}.

\begin{lstlisting}[label=lst:budding-li]
% l_i are the target labels.
% a_i are inputs.
l_1 :- a_0.
l_2 :- a_2, not a_8.
l_2 :- a_1, not a_8.
l_2 :- a_1, a_2.
l_3 :- a_3, not a_8.
l_3 :- a_1, a_3.
l_3 :- a_1, not a_8.
l_4 :- a_2.
l_5 :- not a_4, not a_8, not a_10, a_11..
l_5 :- not a_4, not a_6, not a_8, a_10.
l_5 :- not a_4, not a_6, a_10, a_11.
l_5 :- not a_6, not a_8, a_10, a_11.
l_5 :- a_5, not a_6, not a_8, a_11.
l_5 :- not a_4, a_5, not a_6, not a_8.
l_5 :- not a_4, a_5, not a_8, a_10.
l_5 :- not a_4, a_5, a_10, a_11.
l_5 :- a_5, not a_8, a_10, a_11.
l_5 :- not a_4, a_5, not a_6, a_11.
l_5 :- a_5, not a_6, a_10, a_11.
l_5 :- a_5, not a_6, not a_8, a_10.
l_5 :- not a_4, a_5, not a_6, a_10.
l_6 :- a_3, not a_5, not a_10.
l_6 :- not a_5, a_6, not a_10.
l_6 :- a_3, a_6, not a_10.
l_6 :- a_3, not a_5, a_6.
l_7 :- not a_4, a_7, not a_8, a_10.
l_7 :- not a_4, not a_6, not a_8, a_10.
l_7 :- not a_4, not a_6, a_7, a_10.
l_7 :- not a_4, not a_6, a_7, not a_8.
l_7 :- not a_6, a_7, not a_8, a_10.
l_8 :- not a_5, a_6, not a_7, a_8.
l_8 :- a_6, not a_7, a_9, not a_10.
l_8 :- not a_5, not a_7, a_9, not a_10.
l_8 :- not a_5, a_6, not a_7, a_9.
l_8 :- a_6, not a_7, a_8, a_9.
l_8 :- not a_5, a_6, not a_7, not a_10.
l_8 :- not a_7, a_8, a_9, not a_10.
l_8 :- not a_5, not a_7, a_8, a_9.
l_8 :- a_6, not a_7, a_8, not a_10.
l_8 :- not a_5, not a_7, a_8, not a_10.
l_8 :- not a_5, a_8, a_9, not a_10.
l_8 :- not a_5, a_6, a_9, not a_10.
l_8 :- not a_5, a_6, a_8, not a_10.
l_8 :- not a_5, a_6, a_8, a_9.
l_8 :- a_6, a_8, a_9, not a_10.
l_9 :- a_6.
l_9 :- a_8.
l_10 :- a_8.
l_10 :- a_9.
l_11 :- a_9, a_10.
l_11 :- not a_8, a_9.
l_11 :- not a_8, a_10.
\end{lstlisting}

Below is an example of the logical interpretation of a neural DNF model in the
dataset \textbf{Fission} \citep{bn-fission}.

\begin{lstlisting}[label=lst:bnt-li]
% l_i are the target labels.
% a_i are inputs.
l_1 :- a_0.
l_2 :- a_2, not a_3, a_5, not a_9.
l_2 :- not a_1, not a_3, a_5, not a_9.
l_2 :- not a_1, a_2, not a_3, a_5.
l_2 :- not a_1, a_2, not a_3, not a_9.
l_2 :- not a_1, a_2, a_5, not a_9.
l_3 :- not a_2, a_3, not a_4.
l_3 :- a_3, not a_4, not a_7.
l_3 :- not a_2, a_3, not a_7.
l_3 :- not a_2, not a_4, not a_7.
l_4 :- not a_1, a_4, a_5, not a_9.
l_4 :- not a_1, not a_3, not a_9.
l_4 :- not a_1, not a_3, not a_4, not a_5.
l_4 :- not a_3, a_4, a_5, a_9.
l_5 :- a_7.
l_6 :- a_3, not a_5.
l_6 :- not a_5, a_6.
l_6 :- a_3, a_6.
l_7 :- a_9.
l_8 :- not a_3, a_8.
l_8 :- not a_3, a_5.
l_8 :- a_5, a_8.
l_9 :- not a_2, not a_4, a_6, not a_7, not a_8, a_9.
\end{lstlisting}

Below is an example of the logical interpretation of a neural DNF model in the
dataset \textbf{MAM} \citep{bn-mam}.

\begin{lstlisting}[label=lst:mam-li]
% l_i are the target labels.
% a_i are inputs.
l_0 :- a_0.
l_1 :- not a_2, a_3.
l_2 :- not a_0, a_5, not a_9.
l_2 :- not a_0, not a_1, not a_4, not a_9.
l_3 :- not a_2, a_5, not a_9.
l_3 :- not a_2, not a_4, not a_9.
l_4 :- not a_2, a_3, not a_6, not a_8.
l_4 :- not a_2, a_3, not a_6, not a_7.
l_4 :- not a_2, a_4, not a_6, not a_8.
l_4 :- not a_2, a_4, not a_6, not a_7.
l_5 :- not a_0, not a_4, a_5, not a_9.
l_5 :- not a_0, not a_1, a_5, not a_9.
l_5 :- not a_0, not a_1, not a_4, not a_9.
l_6 :- a_9.
l_7 :- not a_8.
l_7 :- a_7, a_9.
l_7 :- a_6, a_7.
l_7 :- a_4, a_7.
l_8 :- a_6.
l_8 :- a_5, a_9.
l_8 :- not a_4, not a_9.
l_9 :- not a_6, not a_8
\end{lstlisting}

\end{document}